\documentclass{article} 
\usepackage{iclr2027_conference,times}

\usepackage{amsmath,amsfonts,bm}

\def\eqref#1{equation~\ref{#1}}

\def\1{\bm{1}}

\DeclareMathAlphabet{\mathsfit}{\encodingdefault}{\sfdefault}{m}{sl}
\SetMathAlphabet{\mathsfit}{bold}{\encodingdefault}{\sfdefault}{bx}{n}

\usepackage{hyperref}
\usepackage{url}

\usepackage{hyperref}
\usepackage{url}
\usepackage{booktabs}
\usepackage{kotex}
\usepackage{graphicx}
\usepackage{rotating}
\usepackage{amssymb}
\usepackage[mathscr]{euscript}
\usepackage[dvipsnames]{xcolor}
\usepackage{colortbl}
\usepackage{makecell}
\usepackage{subcaption}
\usepackage{multirow}
\usepackage{multicol}
\usepackage{tabularx}
\usepackage{pgfplots}
\usepackage{pgfplotstable}
\usepackage{subcaption}
\usepackage{wrapfig}
\usepackage[most]{tcolorbox}
\usepackage{xcolor}
\usepackage{longtable}
\usepackage{fontawesome5} 
\usepackage[most]{tcolorbox}
\usepackage{listings}
\usepackage{xcolor}

\definecolor{codebg}{RGB}{247,247,247}
\definecolor{codecomment}{RGB}{110,145,120}
\definecolor{codeblue}{RGB}{35,75,145}
\definecolor{codeteal}{RGB}{25,130,130}

\lstdefinestyle{anygoal}{
    basicstyle=\ttfamily\footnotesize\color{black},
    backgroundcolor=\color{codebg},
    commentstyle=\color{codecomment},
    frame=single,
    rulecolor=\color{black!15},
    framerule=0.4pt,
    numbers=none,
    breaklines=true,
    columns=fullflexible,
    keepspaces=true,
    showstringspaces=false,
    morecomment=[l]{\#},
    emph={anygoal},
    emphstyle=\color{codeblue}\bfseries,
    emph={[2]eval},
    emphstyle={[2]\color{codeteal}\bfseries},
}
\definecolor{theoremblue}{RGB}{5,35,90}
\definecolor{theorembg}{RGB}{245,245,250}

\newcounter{propbox}[section]
\renewcommand{\thepropbox}{\thesection.\arabic{propbox}}

\newtcolorbox{propbox}[2]{
  enhanced,
  breakable,
  colback=theorembg,
  colframe=theorembg,
  boxrule=0pt,
  borderline west={2.2pt}{0pt}{theoremblue},
  sharp corners,
  left=8pt,
  right=0pt,
  top=5pt,
  bottom=6pt,
  before skip=8pt,
  after skip=8pt,
  before upper={
    \refstepcounter{propbox}%
    \label{prop:#2}%
    {\bfseries\color{theoremblue} Proposition~\thepropbox}%
    \hspace{0.35em}%
    {\bfseries\color{black} #1}%
    \vspace{1pt}\par
  },
}

\definecolor{notebg}{RGB}{249,249,249}
\definecolor{noteline}{RGB}{150,150,150}

\definecolor{chunksky}{RGB}{241,244,246}

\tcbset{
  notebox/.style={
    enhanced,
    breakable,
    colback=notebg,
    colframe=notebg,
    boxrule=0pt,
    borderline west={1.5pt}{0pt}{noteline},
    sharp corners,
    left=4pt,
    right=4pt,
    top=2pt,
    bottom=2pt,
  }
}

\tcbset{
  bluebox/.style={
    enhanced,
    breakable,
    colback=theorembg,
    colframe=theorembg,
    boxrule=0pt,
    borderline west={1.5pt}{0pt}{theoremblue},
    sharp corners,
    left=4pt,
    right=4pt,
    top=2pt,
    bottom=2pt,
  }
}

\definecolor{pastelyellow}{RGB}{255,248,225}
\definecolor{paleyellow}{RGB}{255,254,250}
\definecolor{darkbrown}{RGB}{70,45,25}

\tcbset{
  interpretationbox/.style={
    enhanced,
    colback=paleyellow,
    colframe=gray!45,
    boxrule=0.6pt,
    arc=2.5mm,
    left=7pt,
    right=7pt,
    top=6pt,
    bottom=6pt,
    fontupper=\small,
  }
}
\pgfplotsset{compat=1.18}
\iclrfinalcopy

\newtheorem{assumption}{Assumption}

\newtheorem{proposition}{Proposition}

\title{
AnyGoal: On World Model Generalization\\  Beyond Expert Rollouts
}

\author{
\resizebox{0.98\textwidth}{!}{%
\hspace{-10pt}
Heejeong Nam$^{1,2}$ \quad
Divyat Mahajan$^{3}$ \quad
Quentin Le Lidec$^{4}$ \quad
Randall Balestriero$^{1}$ \quad
David Klindt$^{2}$} \\[0.35em]
\resizebox{0.97\textwidth}{!}{%
\hspace{-10pt}
$^{1}$Brown University \quad
$^{2}$Cold Spring Harbor Laboratory \quad
$^{3}$Mila, Université de Montréal \quad
$^{4}$New York University\quad
}
}

\begin{document}

\maketitle
\fancyhead{}
\renewcommand{\headrulewidth}{0pt}
\vspace{-1em}
{
\hypersetup{
    colorlinks=false,
    pdfborder={0 0 0}
}
\centerline{\small
  \href{https://hazel-heejeong-nam.github.io/anygoal/}{\faGlobe\enspace Website}\qquad
  \href{https://github.com/galilai-group/AnyGoal}{\faGithub\enspace Code}
}
}
\vspace{1.25em}

\begin{abstract}

The goal of a world model is to capture the dynamics of an environment. This has the great appeal that it allows agents to plan and simulate a course of actions with the help of a world model. For these uses to be reliable, however, the learned dynamics must behave well beyond the specific state–action combinations observed in expert rollouts. 
Despite this, the relationship between data coverage, dynamics generalization, and downstream planning performance remains underexplored. We study this relationship through the lens of compositional generalization: can a world model recombine familiar state and action factors into valid dynamics? To answer this, we introduce AnyGoal, a benchmark of goal-shifted task variants built from widely used world-model environments, together with an open-source library.
Each task is designed to be within reach of a world model that has learned the underlying dynamics: individual state factors and actions remain within their training support, but their combinations are underrepresented in expert trajectories. Across multiple environments and world-model families, we find that current models consistently exhibit failures under these recombinations.
By examining dynamics and representation failures across model families, interaction regimes, and evaluation metrics, we study where current world models generalize and where challenges remain for goal-shifted prediction and planning. Together, this framework, benchmark, and analysis provide a basis for evaluating progress toward world models that generalize to new planning tasks.

\end{abstract}

\vspace{-5pt}
\begin{center}
    \includegraphics[width=\textwidth]{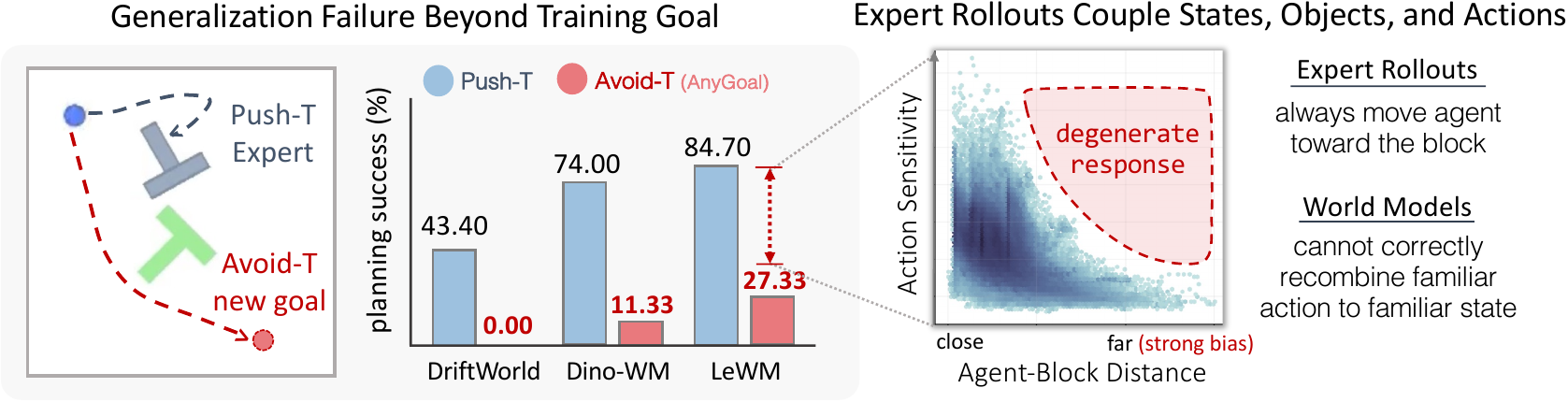}
    \captionof{figure}{\textbf{Example of generalization failure under goal shift.}
A world model trained on Push-T performs poorly on Avoid-T, despite Avoid-T requiring only familiar 2D navigation and having individually familiar states and actions.
In Push-T, action interventions reveal reduced model sensitivity in biased regions, suggesting that the model learns shortcuts rather than reusable dynamics.
}
    \label{fig:teaser}
\end{center}
\vspace{-10pt}

\section{Introduction}

A key goal of physical intelligence is the ability to plan and perform new tasks.
Routine tasks, such as driving to work, can be automatized and executed without much cognitive effort.
Yet novel tasks requires us to plan and simulate the consequences of our actions.
World models~\citep{wm, lewm, sensori-wm, driftworld, irasim, dinowm}
aim to learn action-conditioned dynamics of environment by predicting how the world evolves under actions rather than task-specific behavior~\citep{sutton2018reinforcement}.
Therefore, a good world model should not merely memorize behavioral biases present in the training distribution. However, in-distribution evaluation offers limited evidence of reusable dynamics.
Figure~\ref{fig:teaser} illustrates a simple but strong example of this caveat.
In Push-T, the agent must push a block to a target pose, so expert rollouts strongly couple far-from-block states with actions that move the agent toward the block.
A model trained on Push-T may therefore learn this rollout regularity rather than the action-conditioned effect of the supplied action, as reflected by the action-sensitivity plot in Fig.~\ref{fig:teaser}.When evaluated on Avoid-T, which requires avoiding the block using familiar 2D navigation, the model fails.

If the training data fully covered the relevant joint state--action space, then a model could in principle predict correctly on that support. However, such coverage is unrealistic in many control problems, especially when the relevant space is high-dimensional.
First, offline datasets for world-model training are often collected from goal-directed policies that reach task-relevant states~\citep{burda2018exploration}, which can introduce strong state--action correlations. 
One possible remedy is to mix data from multiple policies~\citep{ICLR2026_7ff013b7}. However, even with such mixtures, it can be difficult to determine whether the resulting dataset sufficiently covers the relevant state--action space, especially when the true reachable set is unknown.
Even when a coverage gap is identified, collecting targeted rollouts to fill it can be difficult in offline settings.
This makes it necessary to study generalization under incomplete coverage, rather than assuming that data scale alone will resolve the issue.

\begin{figure}[t!]
    \centering
    \includegraphics[width=\linewidth]{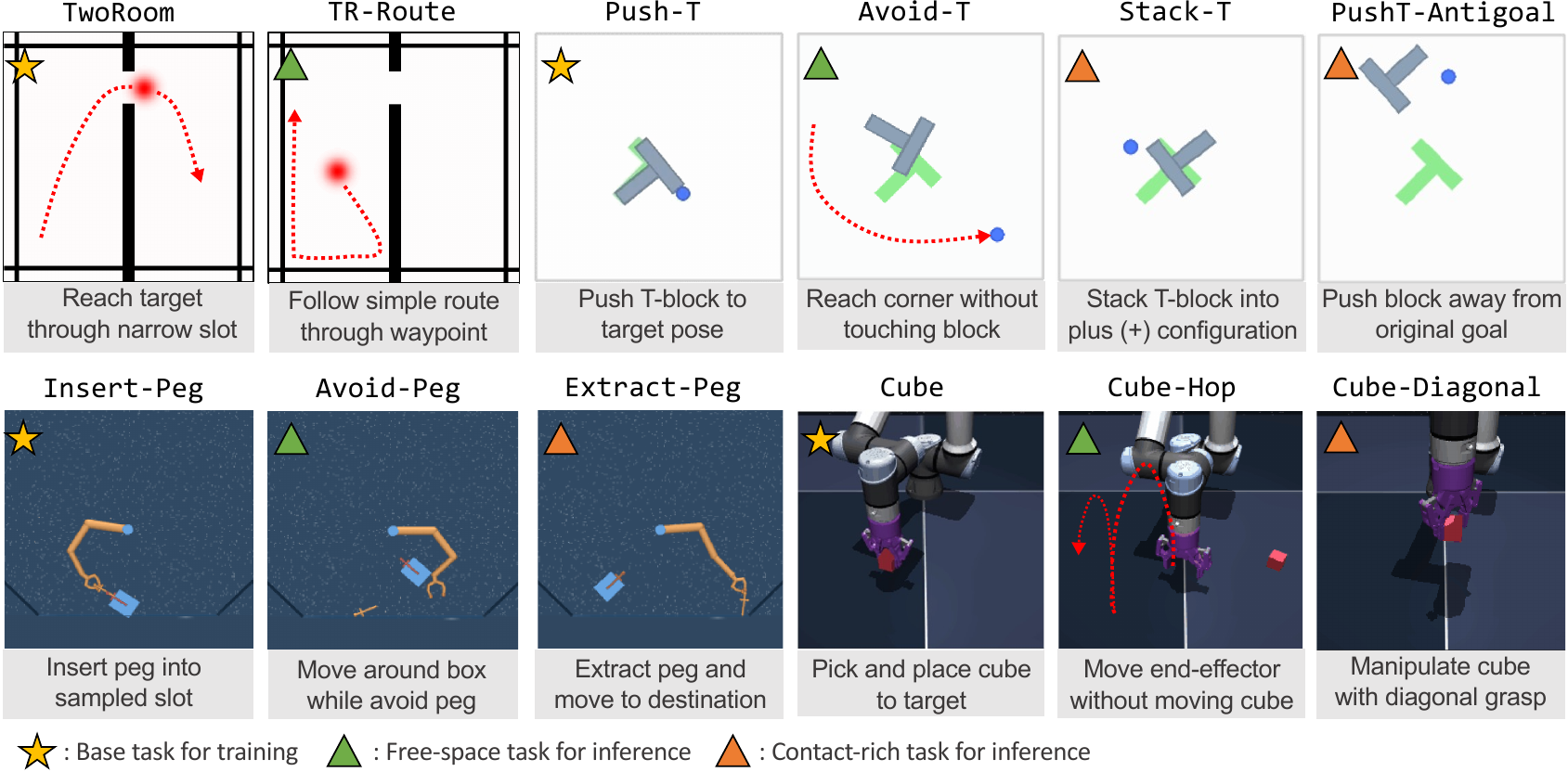}
    \caption{\textbf{Overview of AnyGoal.}
    Each task preserves the environment dynamics of the corresponding base tasks and keep states and actions individually within training coverage.
Throughout the paper, AnyGoal is used solely as a test set for evaluating generalization.}
    \label{fig:benchmark}
\end{figure}

Generalization of world models has been studied to some extent. Closest to our setting, \citet{hansen2026hallucination} study hallucination detection and mitigation in world models, while we evaluate generalization in a statistically controlled space with explicit state-action support checks.
Other OOD evaluations introduce states or factors outside the training support~\citep{swm, adajepa, pusht-m}.
Such settings are valuable for testing robustness to novel factors or extrapolation, but they may go beyond the generalization expected from a reusable dynamics model.

What remains missing is a systematic evaluation of the generalization that a good world model should support.
First, we formalize this criterion as compositional generalization~\citep{wiedemer2024compositional} and link it to goal shifts and behavior bias in the training data. 
We further show that the bias is empirically realized through targeted action interventions on familiar but under-covered state--action combinations, eliciting degenerate model responses.
We then propose AnyGoal (Fig.~\ref{fig:benchmark}), a benchmark of goal-shifted task variants built from widely used world-model environments. To remain in the compositional-generalization regime, AnyGoal preserves the observed marginal support of states, actions, task-relevant factors, and physical rules, while changing the joint combinations induced by the training behavior distribution and targeting under-covered state--action regions.
With AnyGoal, we test whether current world models can generalize to new tasks without requiring extrapolation to unseen states, actions, or visual factors. Across this controlled setting, we observe broad failures across modeling paradigms, including generative world models~\citep{irasim, driftworld} and JEPA-style architectures~\citep{lewm, sensori-wm, dinowm}.
Our results then characterize this degradation across several axes, showing that failures can arise from both dynamics prediction and representation learning, and that their severity depends on task complexity, evaluation metric, and model family.
Finally, we outline a theoretical direction toward compositional generalization for world models, framing what future models may need in order to move beyond behavior-biased rollouts.


\section{Related Works}


\paragraph{Compositional generalization diagnostics and benchmarks.}

Compositional generalization tests whether a model can recombine familiar components into unseen configurations, a key challenge when possible compositions exceed finite data coverage. Early sequence-to-sequence benchmarks such as SCAN~\citep{pmlr-v80-lake18a} and CFQ~\citep{Keysers2020} exposed this failure mode through controlled compositional splits. The same diagnostic was subsequently adopted in vision and multimodal settings, with benchmarks such as CLEVR~\citep{Johnson_2017_CVPR}, Winoground~\citep{Thrush_2022_CVPR}, and CVR~\citep{NEURIPS2022_c08ee8fe} testing whether familiar objects, attributes, and relations can be composed correctly. These failures remain prominent in modern large-scale models: ConceptMix~\citep{wu2024conceptmix} shows that text-to-image models degrade as prompts require more familiar concepts to be composed jointly, while recent vision--language work continues to reveal failures in object--attribute and relational composition~\citep{assouel2026object}. At the same time, compositional evaluation is becoming broader and more systematic, with scalable protocols for probing generalization across controlled compositions~\citep{camposampiero2026scalable}, and similar shifts being used as diagnostics in mechanistic interpretability~\citep{pacela2026stop}. AnyGoal brings this perspective to world models, where goal-directed trajectories naturally induce restricted joint coverage over otherwise familiar physical states and actions.

\paragraph{World-model generalization and support mismatch.}
Recent world-model work has studied generalization through object-centric compositional dynamics~\citep{pmlr-v162-zhao22b}, composable causal components for unseen environments~\citep{ICLR2025_79d86433}, multi-task world models~\citep{ICLR2025_240ea174}, environment-level evaluation queries~\citep{warrier2026benchmarking}, action-conditioned planning~\citep{zhang2026worldactionplannergeneralizable}, and coverage-related hallucination failures~\citep{hansen2026hallucination}. However, less is known about whether expert-trained world models generalize to new goal-induced recombinations of familiar states and actions.
This question is closely connected to support mismatch in offline reinforcement learning, where fixed behavior datasets can lead methods to fail when evaluating poorly covered state--action regions~\citep{fujimoto2019,kumar2019,levine2020offlinereinforcementlearningtutorial}. 
Offline RL addresses support mismatch by constraining or regularizing policies toward covered regions~\citep{gao2025osc,gao2025bdpo,Laroche2019,chen2025active,kumar2020,tang2025robust}. Model-based offline RL extends this issue to learned dynamics, motivating uncertainty- or pessimism-based objectives for poorly covered regions~\citep{yu2020,kidambi2020,yu2021,rigter2022}. D4RL further shows that dataset composition strongly affects offline policy improvement~\citep{fu2020}.
AnyGoal is complementary to these approaches: it evaluates whether the world model learns reusable dynamics under goal shifts.

\section{Compositional Generalization for World Models}
\label{sec:compgen}

When should we say that a world model has learned reusable dynamics, rather than only the training behavior?
In this section, we formalize this criterion as compositional generalization.


\textbf{Notation.} \quad
We use uppercase letters for random variables and lowercase letters for their realizations. Let $z_t = (z_{1,t},\ldots,z_{K,t})$ denote true underlying state of the environment, where each component corresponds to a semantic factor of variation\footnote{We include only factors that vary over the training
distribution. For example, if a block remains gray throughout training, its
color is not included in $z_t$.}~\citep{Scholkopfetal21}, such as object position or end-effector pose. We model observations as \(s_t=f(z_t)\), where \(f\) maps $z_t$ to the RGB observation \(s_t\) presented to the world model. An action-paired trajectory
$
    \tau=(s_0,a_0,s_1,a_1,\ldots,s_H)
$
can therefore equivalently be represented through the corresponding
sequence of physical factors and actions,
$
    (z_0,a_0,z_1,a_1,\ldots,z_H),
$ where \(H\) denotes horizon. 

\subsection{Action-Conditioned Prediction Can Degenerate into Behavior Prediction}
\label{sec:degen}
An action-conditioned world model should learn a transition predictor \(\hat T\) that receives \(s_t\)\footnote{For simplicity, \(s_t\) may represent a finite history of states or observations.} and an action \(a_t\), and predicts how the observation changes under the supplied action, $\hat T(s_t,a_t) \approx s_{t+1}$.
We consider the common offline setting in which data are collected by a behavior policy \(\pi\), so that the observed action satisfies \(a_t \sim \pi(\cdot \mid s_t)\).
Thus, the observed transitions reflect the state--action pairings induced by \(\pi\), rather than arbitrary interventions at each state:
\begin{equation}
T^{\pi}(s_{t+1}\mid s_t)
=
\int
T(s_{t+1}\mid s_t,a)\,
\pi(a\mid s_t)\,da .
\end{equation}
A predictor trained only on such data may degenerate into an action-ignorant model for state regions with restricted conditional action support, fitting the observed transitions by approximating
\begin{equation}
\hat T(s_{t+1}\mid s_t,a_t)
\approx
T^{\pi}(s_{t+1}\mid s_t),
\end{equation}
where the dependence on the supplied action \(a_t\) is effectively lost. 

\subsection{Compositional Distribution Shifts in World Models}
\label{sec:compositional_shift}

Now, let \(g_{\mathrm{tr}}\) and \(g_{\mathrm{te}}\) denote the training and test goals. 
Each goal induces a behavior policy, \(\pi_{\mathrm{tr}}\) or \(\pi_{\mathrm{te}}\), which generates actions along the same environment dynamics \(T\). 
We denote the resulting trajectory distributions by \(p_{\mathrm{tr}}\) and \(p_{\mathrm{te}}\), respectively.
%
We define the supports of physical factors and actions observed across trajectories, writing the training joint state--action support as
\begin{equation}
\mathcal C^{\mathrm{tr}}
:=
\bigcup_{t=0}^{H-1}
\operatorname{supp}_{p_{\mathrm{tr}}}
(Z_{1,t},\ldots,Z_{K,t},A_t).
\end{equation}
We also define the marginal support of each factor as
\(\mathcal Z_k^{\mathrm{tr}} :=
\bigcup_{t=0}^{H-1}
\operatorname{supp}_{p_{\mathrm{tr}}}(Z_{k,t})\), and the action support as
\(\mathcal A^{\mathrm{tr}} :=
\bigcup_{t=0}^{H-1}
\operatorname{supp}_{p_{\mathrm{tr}}}(A_t)\).
These supports pool over temporal position: a physical factor is considered
observed during training if it occurs at any point along a trajectory drawn
from $p_{\mathrm{tr}}$.
The test supports \(\mathcal C^{\mathrm{te}}\), \(\mathcal Z_k^{\mathrm{te}}\), and \(\mathcal A^{\mathrm{te}}\) are defined analogously under \(p_{\mathrm{te}}\).
We define the \emph{combinatorial space} as the Cartesian product of the marginal training supports:
\begin{equation}
    \mathcal C^\times
    :=
    \left(
        \prod_{k=1}^{K}\mathcal Z_k^{\mathrm{tr}}
    \right)
    \times
    \mathcal A^{\mathrm{tr}}.
\end{equation}
\begin{wrapfigure}{l}{0.30\linewidth}
    \centering
    \vspace{-2em}
    \includegraphics[width=\linewidth]{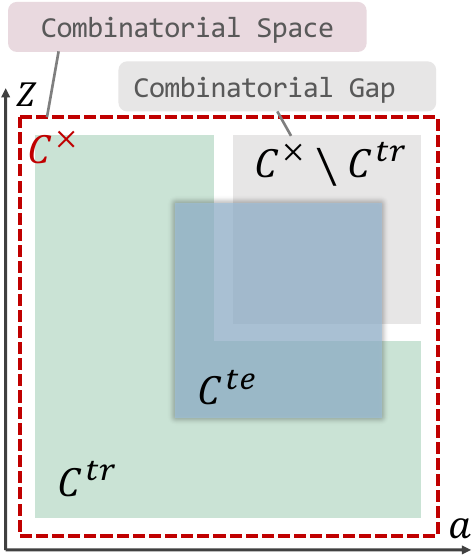}
    \caption{Compositional support structure. $\mathcal C^{\mathrm{te}}
    \subseteq
    \mathcal C^\times$ and $C^{te}\not \subset \mathcal C^{\mathrm{tr}}$is necessary.}
    \label{fig:analysis}
    \vspace{-1em}
\end{wrapfigure}

We study compositional distribution shifts~\citep{liu2023causal, mahajan2025compositional} where the underlying transition dynamics are invariant across the train and test distribution, i.e., $T_{{\mathrm{tr}}}(s_{t+1}\mid s_t,a_t) = T_{{\mathrm{te}}}(s_{t+1}\mid s_t,a_t)$, but the test support introduces state, action pairs not present in the training support ($\mathcal C^{\mathrm{te}}  \not \subset \mathcal C^{\mathrm{tr}} $). AnyGoal targets the \emph{combinatorial gap}, regions of the combinatorial space \(\mathcal C^\times\) that are absent from the training joint support \(\mathcal C^{\mathrm{tr}}\). Since test trajectories need not cover the entire gap or lie only within it, we require \(\mathcal C^{\mathrm{te}} \subseteq \mathcal C^\times\) and choose goals with substantial mass in under-covered regions of \(\mathcal C^\times\).
This means that the test distribution does not introduce novel physical factors or actions, only novel combinations of observed physical factors and actions. 
%
%
Importantly, training-distribution accuracy does not identify the extension of the learned transition function to \(\mathcal C^\times \setminus \mathcal C^{\mathrm{tr}}\): predictors that agree on the training support may disagree on novel compositions~\citep{wiedemer2024compositional}. This motivates evaluating the combinatorial gap directly. We discuss more about why it is an essential test in Appendix~\ref{app:compositional-world-model-test}.


\section{AnyGoal: Controlled Goal-Shift Evaluation}
\label{sec:benchmark}

We introduce AnyGoal, a benchmark for testing whether world models generalize beyond the goal-directed trajectories on which they are trained. AnyGoal uses the controlled goal shifts discussed in Sec.~\ref{sec:compositional_shift} and builds on four widely used world-model tasks: TwoRoom~\citep{pldm}, Push-T~\citep{pusht}, DMC Manipulator Insert-Peg~\citep{tassa2018deepmindcontrolsuite}, and OGBench Cube~\citep{ogbench}. 
AnyGoal contains eight variants: one free-space variant per base task and one or two contact-rich variants except TwoRoom. We release 400 episodes per variant, with data-collection tools for generating additional datasets. Figure~\ref{fig:benchmark} summarizes the tasks, with details in Appendix~\ref{app:data_detail}.


\begin{figure}[t!]
    \centering
    \includegraphics[width=1\linewidth]{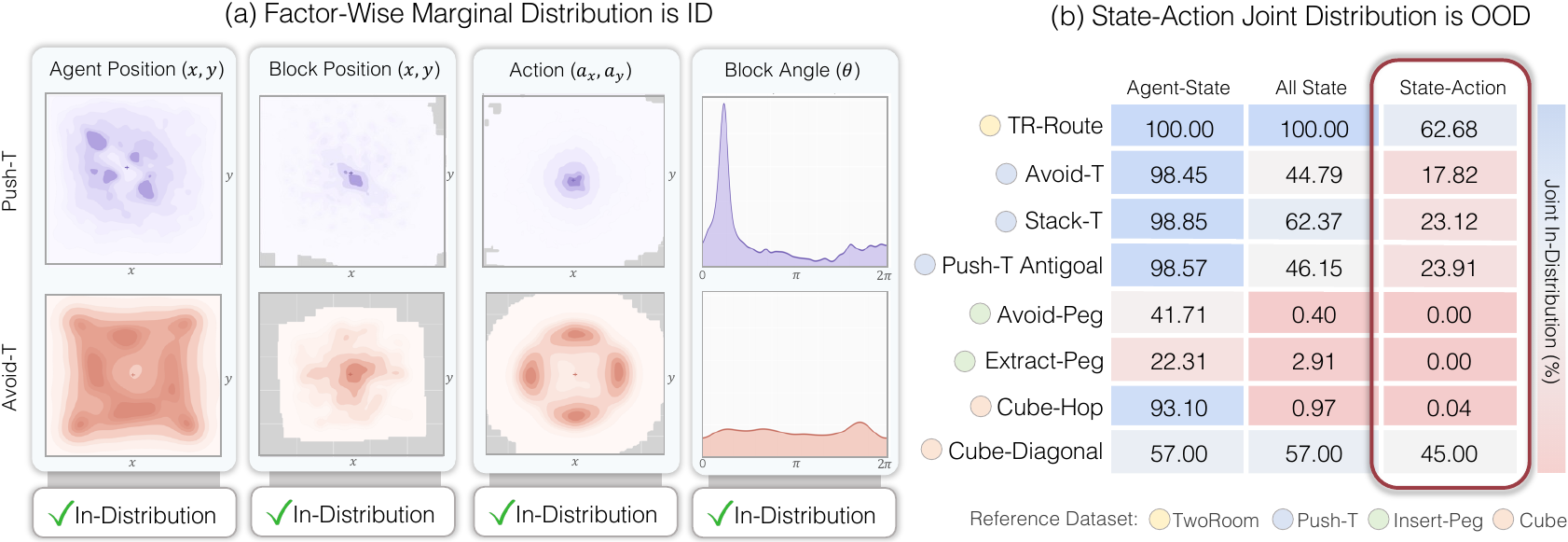}
\caption{
\textbf{AnyGoal targets the combinatorial gap.}
(a) Example factor-wise support check for Avoid-T relative to Push-T using KDE. Darker regions indicate higher density, and gray regions mark areas whose density is below \(1\%\) of the peak reference density. 
(b) Estimated overlap between each target variant and the base-task training distribution via KDE.
}
    \label{fig:placeholder}
\end{figure}

\subsection{AnyGoal Targets the Combinatorial Gap}
\label{sec:support_verification}


\textbf{\boldmath Validating $\mathcal C^{\mathrm{te}} \subseteq \mathcal C^\times$.} \quad For each base--target task pair, we apply three factor-wise support checks to verify that the target distribution remains marginally in-distribution relative to its reference distribution. The range check requires all query samples to lie within the coordinate-wise reference range. The Mahalanobis check uses the reference mean and covariance, thresholds distances at the reference \(0.99\)-quantile, and requires at least \(95\%\) of query samples to fall below it. The KDE check thresholds reference densities at the \(0.01\)-quantile and requires at least \(95\%\) of query samples to lie above it. The final support verdict is the conjunction of all three checks.
All AnyGoal variants are designed to pass these factor-wise checks, which indicates that all tasks lie within the combinatorial space of training data.
Appendix~\ref{app:distribution_support_criteria} provides the full details of pass-criterion definitions, the verified factor groups for each base task, and the complete score of range, Mahalanobis, and KDE results.

\textbf{\boldmath Measuring combinatorial-gap coverage of $\mathcal C^{\mathrm{te}}$.} \quad
We then estimate how much each target distribution overlaps with the base-task joint support. Unlike the factor-wise checks above, this joint test is meant to diagnose the combinatorial gap: individual factors may be marginally familiar while their full state--action combination is rare under the training distribution. For each base--target pair, we concatenate the action, agent state, and task-object or target state, fit a full-dimensional Gaussian KDE to the reference vectors, and set the density threshold to the empirical \(0.01\)-quantile of log-densities.
We then report the fraction of target vectors whose log-density exceeds this threshold.
Lower joint scores indicate less overlap with the base-task joint distribution and greater coverage of the combinatorial gap. Full vector definitions and KDE implementation details are in Appendix~\ref{app:joint_distribution_support}.

\subsection{Relative Evaluation Against Base Tasks}
\label{sec:evaluation_protocol}

We evaluate rollout prediction error and downstream MPC planning. Models are trained only on the base dataset and evaluated on both base and goal-shifted trajectories to measure relative changes. Rollout-error changes more directly diagnose whether action-conditioned dynamics generalize to unseen state--action combinations.
However, planning success remains important. For JEPA-style models, it provides complementary evidence that the representation retains task-relevant information. For generative models, low rollout error can miss small but task-critical observation changes. We therefore report both metrics over seeds \(0,1,2\). Evaluation details are in Appendix~\ref{app:evaluation}

\textbf{Prediction Error.}
We measure MSE in each model's native prediction space, which may be a learned latent space, oracle state space, or pixel space depending on the model class.
We evaluate autoregressive prediction over horizons of 5 and 25 environment steps, with frame skip of 5. For each seed, we sample 100 episodes and average over all valid starting points within each episode.

\textbf{MPC Planning Success Rate.}
We evaluate downstream control using the same CEM-based MPC implementation from Stable-WorldModel~\citep{swm} for all models. We use identical planning configurations, with a planning horizon of 25 environment steps. The full 25-step action sequence is executed before replanning.

\subsection{AnyGoal Library and Evaluation Interface}
\label{sec:anygoal_library}

We release \texttt{AnyGoal} as an open-source Python library for reproducible
goal-shift evaluation. The library provides dataset access, task-specific
data collection, and standardized rollout-prediction and MPC-planning
evaluation across supported world models. Its unified CLI allows
researchers to evaluate the same base-trained model on both the original
task and unseen goals, with explicit model, dataset, and seed selection.
The library also exposes Python APIs for dataset collection, model
training, and evaluation. We provide examples and reproducibility
instructions in Appendix~\ref{app:anygoal_library}, with complete
documentation available \href{https://hazel-heejeong-nam.github.io/anygoal/docs/}{online}.

\begin{figure}[h]
\centering
\begin{minipage}{0.9\linewidth}
\begin{lstlisting}[style=anygoal]
# Original training goal
anygoal eval planning --model lewm --dataset pusht --seed 0

# Unseen goal
anygoal eval planning --model lewm --dataset avoidt --seed 0
\end{lstlisting}
\end{minipage}

\caption{
The same base-trained model is evaluated on Push-T and Avoid-T using
the standardized planning interface.
}
\label{fig:anygoal_cli}
\end{figure}







\section{Experiment and Results}
\label{sec:experiments}


\begin{table}[t]
\centering
\small
\setlength{\tabcolsep}{3pt}
\renewcommand{\arraystretch}{0.9}
\caption{End-to-End World models and oracle state predictors evaluated in our experiments.}
\label{tab:evaluated_models}
\begin{tabularx}{\linewidth}{
    >{\raggedright\arraybackslash}p{0.25\linewidth}
    >{\raggedright\arraybackslash}p{0.20\linewidth}
    >{\raggedright\arraybackslash}p{0.22\linewidth}
    >{\raggedright\arraybackslash}X}
\toprule
Model & Family & Representation & Action conditioning \\
\midrule
\multicolumn{4}{l}{\textit{\textbf{End-to-End World Models}}} \\
\addlinespace[2pt]

LeWorldModel
& JEPA
& Learned latent
& AdaLN \\

Sensorimotor WM
& JEPA
& Learned latent
& AdaLN \\

DINO-WM
& JEPA
& Frozen DINO features
& Feature concatenation \\

DriftWorld
& Generative
& Pixels
& Conditional generator \\

IRASim
& Generative
& Frozen SDXL VAE
& DiT conditioning \\

\midrule
\multicolumn{4}{l}{\textit{\textbf{Predictor-Only Models}}} \\
\addlinespace[2pt]

AdaLN
& Transformer
& Oracle state
& AdaLN \\

Action-Concatenated
& Transformer
& Oracle state
& Feature concatenation \\

Toy-DriftWorld
& Generative
& Oracle state
& Conditional generator \\

\bottomrule
\end{tabularx}
\end{table}

We evaluate a diverse model suite spanning JEPA-style and generative objectives, learned and frozen visual encoders, and action-conditioning mechanisms (Tab.~\ref{tab:evaluated_models}). To separate dynamics learning from visual representation failure, we also include oracle-state predictors which replace visual encoders with oracle states. For consistency, all models use frame skip \(5\) and one-step prediction. Model details and oracle-state variables for each environment are provided in Appendix~\ref{app:model_details}, and the full numerical results are summarized in Appendix~\ref{app:full-results}.

\textbf{End-to-End World Models.} \quad
LeWorldModel (LeWM)~\citep{lewm} is an end-to-end JEPA that jointly learns a visual encoder and an action-conditioned transformer predictor from pixels. It uses SIGReg~\citep{lejepa} for anti-collapse regularization and conditions the predictor through AdaLN~\citep{Peebles_2023_ICCV}.
Sensorimotor World Model (SMWM)~\citep{sensori-wm} uses the same encoder--predictor family as LeWM with an inverse-dynamics objective that encourages representations to retain controllable changes.
DINO-WM~\citep{dinowm} decouples perception from dynamics by using a frozen DINOv2~\citep{dinov2} patch encoder and predicting future patch representations, with actions concatenated to the patch tokens.
DriftWorld~\citep{driftworld} is a generative model that uses an action-conditioned spatial--temporal U-Net to predict future observations directly from observation history, actions, and noise.
IRASim~\citep{irasim} is a latent generative model that encodes observations with a frozen pretrained VAE and uses a Diffusion Transformer~\citep{Peebles_2023_ICCV} to predict future latents conditioned on action trajectories.

\vspace{2pt}
\textbf{Oracle-State Predictor.} \quad
AdaLN predictor follows the architecture used by LeWM and SMWM.
Action-Concatenated predictor follows the DINO-WM predictor, where an embedded action is concatenated to the patch representation before being processed by a causal transformer.
Toy-DriftWorld is adapted from DriftWorld's action-conditioned generative backbone. We use a reduced DriftWorld-style model and replace pixel inputs and outputs with true state inputs and next-state targets.



\subsection{Dimensions of Goal-Shift Performance}

Goal-shift performance cannot be summarized by a single number. The same degradation can reflect different sources depending on whether it appears in prediction or planning, in latent or pixel space, in free-space or contact-rich tasks, and with or without oracle state access. We therefore organize the results around five recurring dimensions for interpreting the results:

\begin{tcolorbox}[interpretationbox]
{\textbf{\textcolor{darkbrown}{1. Failure source}}: } 
Predictor failures can be isolated from encoder failures via oracle-state predictors.

\vspace{3pt}
{\textbf{\textcolor{darkbrown}{2. Evaluation metric}}: }
Rollout prediction and planning are showing different aspects.

\vspace{3pt}
{\textbf{\textcolor{darkbrown}{3. Task complexity}}: } 
Free-space and contact-rich tasks impose different dynamics challenges.

\vspace{3pt}
{\textbf{\textcolor{darkbrown}{4. Representation quality}}: } 
Learned representations sometimes obscure task-relevant dynamics, although they sometimes can improve planning geometry.

\vspace{3pt}
{\textbf{\textcolor{darkbrown}{5. Prediction space}}: } 
JEPA-style and generative models show distinct trends, but comparisons should be interpreted cautiously because prediction errors are measured in different spaces.
\end{tcolorbox}

\subsection{Dynamics Generalization Beyond Representation Failure}
\begin{figure}[t!]
    \centering
    \includegraphics[width=1\linewidth]{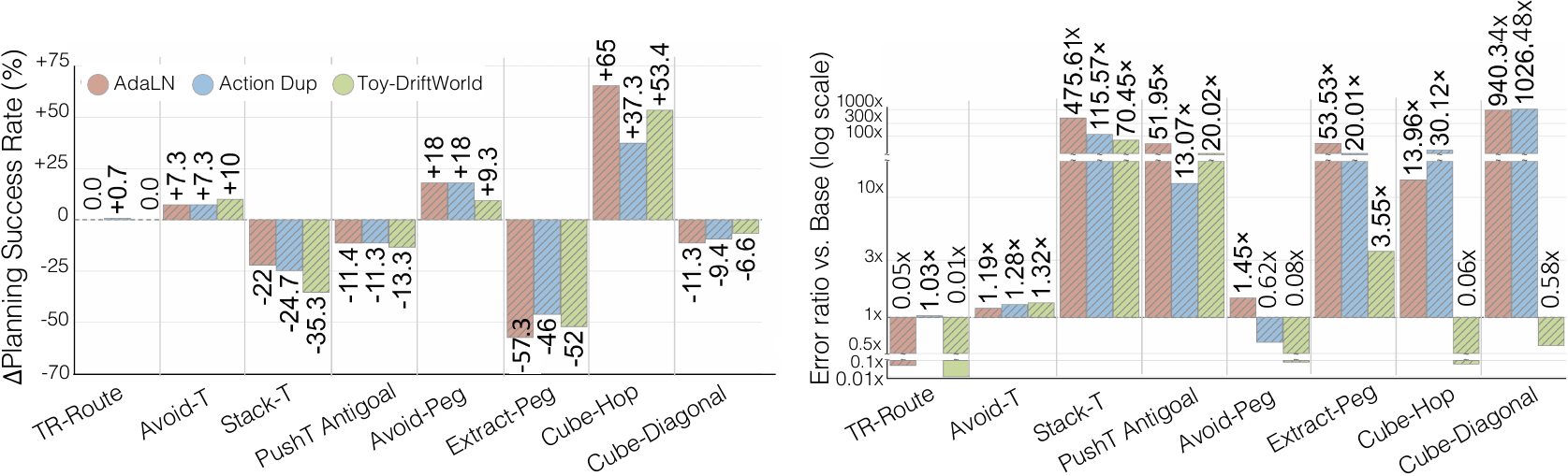}
    \caption{
\textbf{Oracle-state predictors reveal dynamics-level failures.}
Even with task-relevant state factors provided directly, goal shifts still degrade prediction and planning, showing that the AnyGoal gap cannot be reduced to visual representation failure.  See Appendix~\ref{app:full-results}for absolute metrics.
}
    \label{fig:toy}
\end{figure}
To examine dynamics generalization, we use the oracle states provided by AnyGoal, which expose task-relevant factors for action-conditioned next-step prediction and remove the need to infer them from observations under goal shifts. Overall, oracle-state predictors show that the AnyGoal gap cannot be reduced to visual representation failure.

\textbf{Oracle-state predictors reveal an interaction-dependent dynamics gap.} \quad Figure~\ref{fig:toy} shows that removing visual abstraction does not yield a uniform outcome across AnyGoal variants. In free-space variants such as TR-Route, Avoid-T, Avoid-Peg, and Cube-Hop, MPC performance is preserved or improves relative to the corresponding base task. In contrast, contact-rich variants consistently degrade, despite using the same privileged state inputs. This suggests that the difficulty is not merely access to the relevant state factors \citep{klindt2026doeslejepalearnworld, shi2026overcomingstatisticalbiasactioncontrollable}, but whether the goal shift requires precise object interaction.  
This asymmetry is consistent with prior work showing that compositional generalization requires constraints on the learned function. Additive or interaction-free settings are especially favorable, since factor-specific mechanisms can be learned from observed compositions and reused under novel recombinations~\citep{dong2022first, lachapelle2023additive, wiedemer2024provable, mahajan2025compositional}. 
This intuition aligns with the independent causal mechanisms principle, under which reusable mechanisms remain autonomous as their joint occurrence changes~\citep{parascandolo2018learning}. Contact-rich variants depart from this favorable regime: action effects may no longer decompose into reusable factor-specific mechanisms. Compositional generalization may still be possible, but only with additional structure on these interactions~\citep{brady2025interaction}.
%
The same asymmetry appears in rollout prediction. Contact-rich variants show much larger error increases, often by one to three orders of magnitude, even with oracle states. Free-space variants degrade far less, and in some cases match or improve on the base-task error. Thus, removing representation learning does not make goal-shifted dynamics prediction uniformly reliable.

\textbf{Metric alignment depends on task complexity.} \quad We additionally observe that rollout error and planning performance sometimes diverge in free-space variants where planning can tolerate imperfect prediction. The closer agreement in contact-rich settings may reflect the structure of the planning problem. Rollout error measures whether the model predicts the full trajectory accurately, whereas MPC only needs to find some action sequence that reaches the goal, regardless of the exact intermediate path. 
Free-space tasks often admit many such trajectories, while contact-rich tasks allow fewer viable sequences because success also depends on the correct object interaction.


\subsection{Representation Effects in End-to-End World Models}
End-to-end world models can degrade on both metric under the goal shift even when oracle-state predictors remain stable. Such cases show that the AnyGoal gap is not solely a failure of action-conditioned dynamics, but can also reflect encoder failure under unseen factor combinations. Due to space constraints, we discuss two main observations, and more are in Appendix~\ref{app:additional-observations}.

\begin{figure}[t!]
    \centering
    \includegraphics[width=1\linewidth]{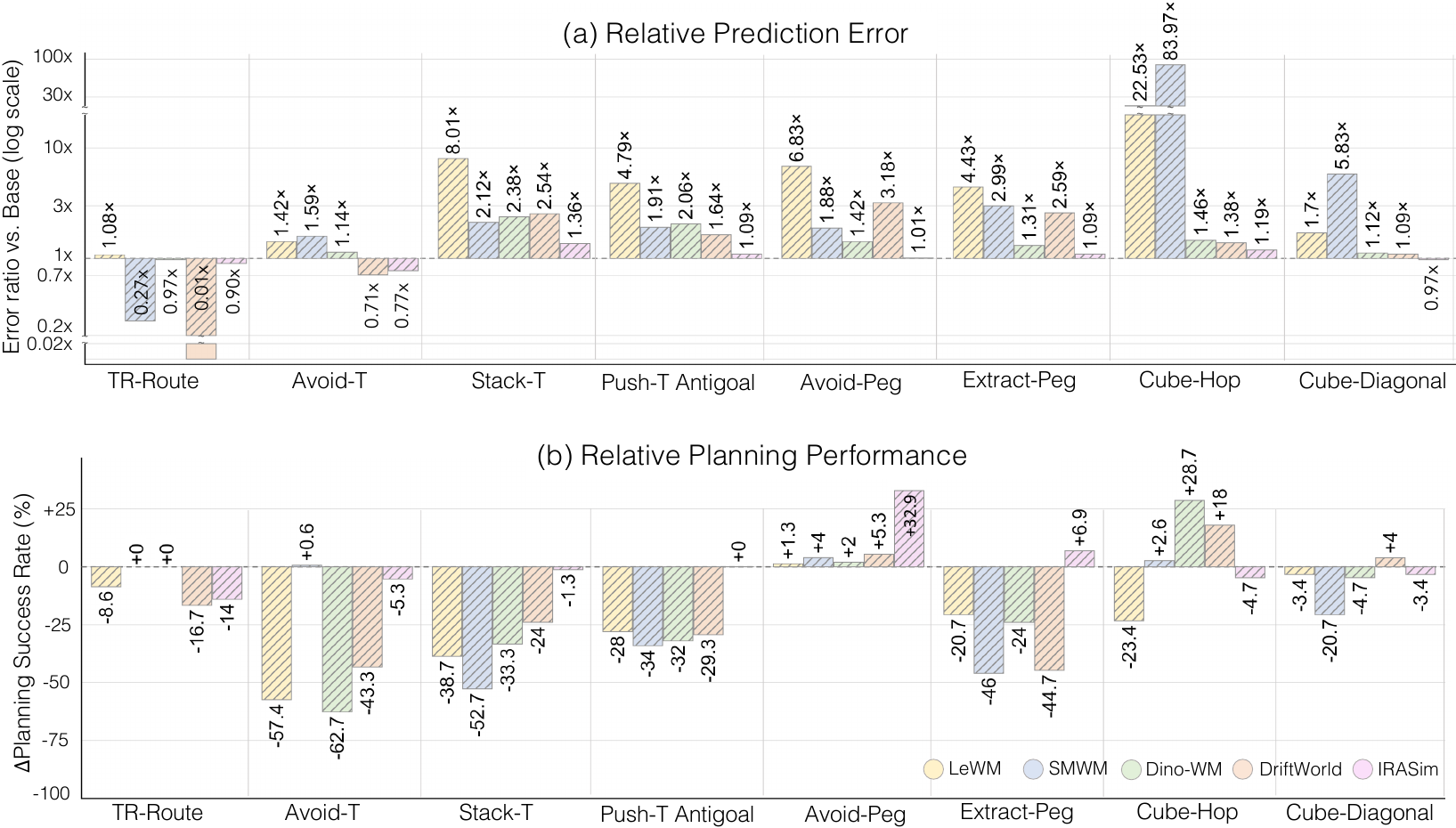}
    \caption{
\textbf{End-to-end degradation reflects representation effects.}
Compared with oracle-state predictors, end-to-end models degrade more under goal shifts, suggesting additional failures from observation-level shift and learned representation geometry. See Appendix~\ref{app:full-results}for absolute metrics.
}
    \label{fig:e2e_planning}
\end{figure}


\textbf{Factor recombination can induce observation-level shifts.} \quad
AnyGoal controls generalization at the level of physical factors, but end-to-end models observe rendered inputs \(s=f(z)\). Recombining marginally familiar factors can therefore create observations outside the joint visual support of expert rollouts, adding an encoder-inference challenge beyond dynamics prediction. (See Appendix~\ref{app:factor_vs_obs_support} for details.)
Crucially, correct factor inference on training observations need not extend to unseen compositions. Even when factors are identifiable in distribution, an encoder may fail to invert novel recombined observations without compositional-consistency constraints~\citep{wiedemer2024provable}. Thus, end-to-end world models face a representation-level generalization problem before dynamics prediction is considered.
End-to-end models must therefore infer task-relevant factors from observations generated by unseen factor combinations, in addition to learning dynamics.

\textbf{Analysis of goal shifts without observation-level shift.} \quad
\begin{figure}[t!]
    \centering
    \includegraphics[width=\linewidth]{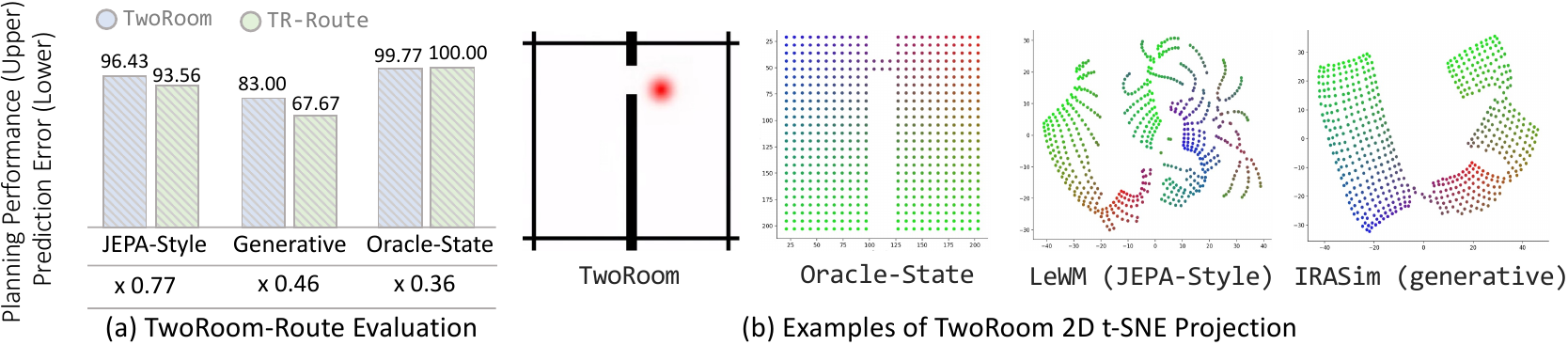}
    \caption{\textbf{TwoRoom as a Visually In-Support Goal-Shift Case Study.}
(a) Rollout prediction improves across model families, while planning degrades for end-to-end models and remains stable for oracle-state predictors. (b) The 2D t-SNE projections illustrate that learned representations can organize the same underlying agent positions with substantially different geometry, which can affect prediction and planning.
}
    \label{fig:tworoom}
\end{figure}
TwoRoom provides a useful case where end-to-end degradation cannot be attributed to an observation-level shift. 
Because agent position is the only changing factor and TR-Route positions remain marginally within the TwoRoom training support, the corresponding observations stay visually in support, enabling a controlled comparison between end-to-end models and oracle-state predictors.
%
Figure~\ref{fig:tworoom} (a) summarizes results from TwoRoom and TR-Route across different model families. We observed that rollout prediction improves across model families, while planning degrades for end-to-end models and remains stable for oracle-state predictors. This suggests that lower prediction error does not necessarily translate into better control: the learned latent space may still distort task-relevant state differences or action effects needed for planning. Thus, even visually in-support goal shifts can expose a mismatch between predictive accuracy and planning utility.

\subsection{Diagnosing Behavior-Policy Shortcuts Through Action Sensitivity}
\label{sec:action_sensitivity}

As discussed in Sec.~\ref{sec:degen}, a world model trained on expert rollouts may exploit correlations between states and the actions typically taken by the expert, rather than learning the effects of the supplied action.
We investigate whether this failure mode is reflected in the learned models by measuring their sensitivity to interventional action inputs.

\textbf{Interventional action sensitivity.}
To test whether the model responds to the supplied action rather than relying on the behavior policy's preferred transition, we intervene
on the action input while holding the state fixed. For each training state $s$ with observed action $a$, we replace $a$
with actions from a fixed interventional set $\mathcal{A}_{\mathrm{cf}}$ and measure the resulting change in the model's predicted representation:
\begin{equation}
\label{eq:action_sensitivity}
S_\theta(s)
=
\frac{1}{|\mathcal{A}_{\mathrm{cf}}|}
\sum_{a' \in \mathcal{A}_{\mathrm{cf}}}
\left\|
\phi_\theta(s,a)
-
\phi_\theta(s,a')
\right\|_2 ,
\end{equation}
where $\phi_\theta(s,a)$ denotes the model representation used for
the sensitivity measurement. A smaller score indicates that the model is less sensitive to changes in the supplied action, although this does not necessarily imply incorrect predictions.
We evaluate this score across training states and relate it to an environment-specific coordinate along which expert behavior is
strongly biased. More details are in
Appendix~\ref{app:sensitivity}.

\begin{figure}[t]
    \centering
    \includegraphics[width=1\linewidth]{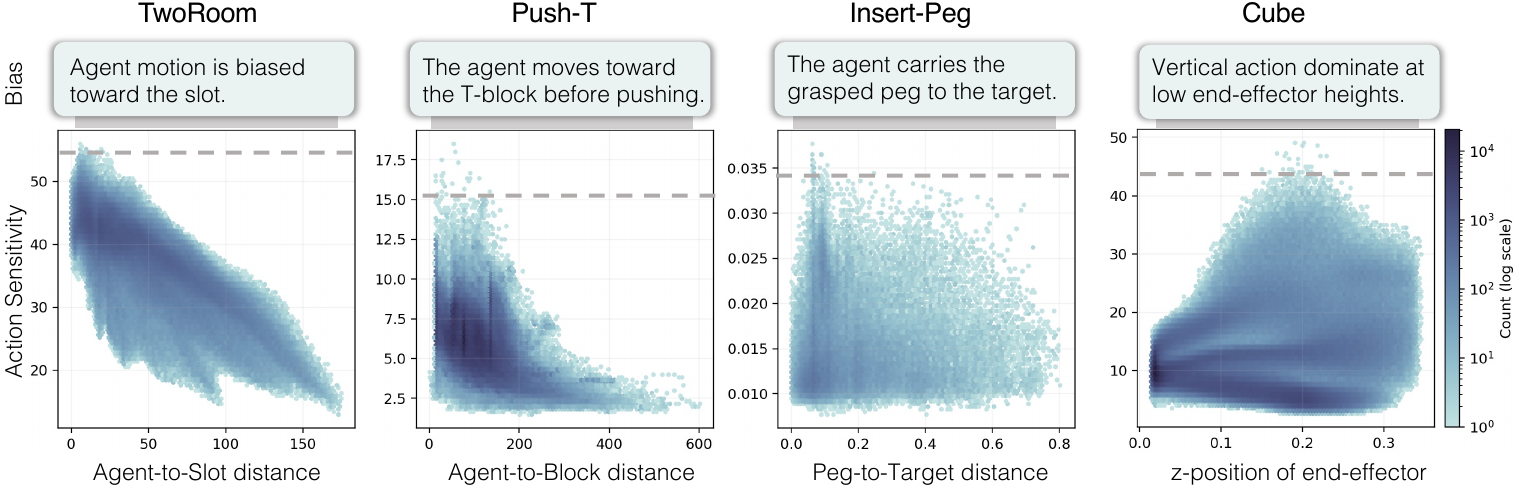}
    \caption{
    \textbf{Goal-biased training support is associated with reduced action sensitivity.}
    Each panel plots model action sensitivity against an environment-specific bias coordinate induced by the expert policy. The analysis suggests that the model can partially rely on a goal-directed prior rather than the supplied action.
    }
    \label{fig:bias}
\end{figure}

\textbf{Goal bias and reduced action sensitivity.}
Figure~\ref{fig:bias} shows that action sensitivity varies systematically with the state regions associated with expert
behavior. In TwoRoom, expert trajectories favor motion toward the narrow passage. In Push-T, the agent typically approaches
the block before pushing it. In Insert-Peg, the agent carries the grasped peg toward the target and in Cube, vertical actions
are prevalent at low end-effector heights. Across these environments, we observe reduced sensitivity to
interventional actions in regions where the training behavior is strongly constrained by the goal and these observations are consistent with the degeneracy described in Sec.~\ref{sec:degen}.

\section{Towards world models that generalize compositionally}

Our oracle-state experiments show that compositional failures cannot be attributed only to visual representation learning. Even when task-relevant physical state is provided directly, goal shifts remain difficult, particularly when prediction involves contact-rich interactions. This raises a broader question for future work: what assumptions on the dynamics, data, and learned representation are sufficient for a world model to generalize compositionally? We see two distinct challenges. 

\begin{tcolorbox}[notebox]
\textbf{1. Identifying mechanisms.} \quad Fitting training trajectories does not ensure that a predictor has learned reusable action-conditioned mechanisms. This becomes harder in interaction-rich dynamics where action effects can vary with state and regime. Characterizing when such mechanisms can be identified from offline trajectories remains an open problem.
\vspace{8pt}

\textbf{2. Extrapolation beyond the training support.} \quad Even if the correct mechanisms are identified, their behavior on novel state--action compositions is generally not determined without additional structural assumptions. Interaction-free dynamics provide a favorable case for compositional generalization because factor effects can often be reused independently, but world models must also handle interaction-rich dynamics where action effects vary with the joint physical configuration and regime. In this setting, extrapolation is fundamentally harder because the relevant interactions may change across compositions, and is a fruitful direction for future work.
\end{tcolorbox}

\textbf{A simple illustrative case.}\quad
One favorable case for extrapolation in interaction-rich dynamics arises when, within each interaction regime, each transition mechanism depends only on a small subset of the state and action variables. Consider regime-dependent dynamics of the following form $z_{t+1}=T_{r_t}(z_t,a_t)$,
where $r_t$ denotes an interaction regime, such as free motion, contact, or grasping. Both the autonomous state evolution and the effect of the action may therefore change across regimes. Suppose that, within a fixed regime $r$, the transition of each next-state factor $z_{k,t+1}$ depends only on a subset of the coordinates of $(z_t,a_t)$. Writing \(x_t=(z_t,a_t)\), this sparse dependence can be expressed as follows, $\left\|\nabla_x T_{k, r_t}(x)\right\|_0 \ll \dim(x)$. 

Let \(u_{k}\) denote the relevant state and action coordinates for $k$-th transition mechanism $T_{k,r_t}$, so that we may write $z_{k,t+1}=T_{k,r_t}(u_{k, t})$. This allows a simple case where compositional shift in the full state--action space need not require extrapolation for each transition mechanism. 
\[
    \mathcal C^{\mathrm{te}}\not\subseteq\mathcal C^{\mathrm{tr}},
    \qquad\text{while}\qquad
    \mathcal U_{k}^{\mathrm{te}}
    \subseteq
    \mathcal U_{k}^{\mathrm{tr}}
    \quad \forall\, k
\]
where \(\mathcal U_{k}^{\mathrm{tr}}\) and \(\mathcal U_{k}^{\mathrm{te}}\) denote the train and test supports of \(u_{k}\) for the fixed regime \(r\). Thus, the complete state--action composition may be unseen, while every input relevant to predicting each next-state factor remains within its training support. Hence, if the corresponding mechanisms have been correctly learned, we can generalize to novel compositions as we do not need to extrapolate any individual mechanism beyond its training support. Appendix~\ref{app:future_compgen_theory} (Proposition~\ref{prop:local_coverage_error}) gives a formal statement.

\section{Conclusion}

We introduced AnyGoal, a benchmark for evaluating whether world models generalize to new behavioral goals by recombining familiar states and actions. Across a diverse set of models, goal-shift generalization remains difficult even when the relevant state and action factors are marginally covered by the training distribution.
%
Together, these analyses show that goal-shift failures are heterogeneous, reflecting interactions among representation, dynamics prediction, task structure, and evaluation metric.
More broadly, AnyGoal connects world-model evaluation with compositional generalization. It asks for prediction on novel joint state--action combinations within the product of marginal training supports, a reasonable but unresolved requirement for reusable dynamics. Our theoretical perspective suggests that global coverage of all state--action combinations is neither realistic nor necessary. Instead, world models need coverage of local mechanism inputs, together with objectives or architectures that identify action-conditioned mechanisms rather than behavior-policy shortcuts. By isolating goal-induced recombination as a failure mode, AnyGoal provides a controlled basis for evaluating progress toward world models that learn reusable dynamics beyond expert rollouts.

\subsubsection*{Acknowledgments}
David Klindt acknowledges the CSHL GPU cluster with assistance from the US National Institutes of Health Grant S10OD028632-01. Divyat Mahajan acknowledges support via FRQNT doctoral scholarship (\url{https://doi.org/10.69777/354785}) for his graduate studies.

\bibliography{iclr2027_conference}
\bibliographystyle{iclr2027_conference}

\appendix

\newpage
\begingroup
\setlength{\parindent}{0pt}
\setlength{\parskip}{0pt}

\begin{center}
{\Large Supplementary Material\par}
\vspace{10pt}
{\LARGE AnyGoal: On World Model Generalization \\Beyond Expert Rollouts\par}
\end{center}
\vspace{2.25em}

\newcommand{\appTOCsection}[3]{%
  \noindent\textbf{#1\hspace{0.45em}#2}\nobreak
  \leaders\hbox{\kern0.25em.\kern0.25em}\hfill
  \nobreak\makebox[2.2em][r]{\pageref{#3}}\par
  \vspace{0.65em}%
}
\newcommand{\appTOCsubsection}[3]{%
  \noindent\hspace{1.5em}#1\hspace{0.45em}#2\nobreak
  \leaders\hbox{\kern0.25em.\kern0.25em}\hfill
  \nobreak\makebox[2.2em][r]{\pageref{#3}}\par
  \vspace{0.4em}%
}

\appTOCsection{A.}{Compositional Generalization: Essential World-Model Test}{app:compositional-world-model-test}
\appTOCsection{B.}{AnyGoal Task Details}{app:data_detail}
\appTOCsubsection{B.1}{Two Room and Its Variants}{app:two-room-variants}
\appTOCsubsection{B.2}{PushT and Its Variants}{app:pusht-variants}
\appTOCsubsection{B.3}{Insert-Peg and Its Variants}{app:insert-peg-variants}
\appTOCsubsection{B.4}{OGBench Cube and Its Variants}{app:ogbench-cube-variants}
\appTOCsection{C.}{Statistical Test 1: Marginal Distribution Support}{app:distribution_support_criteria}
\appTOCsubsection{C.1}{Distribution Support Criteria}{app:distribution-support-criteria-details}
\appTOCsubsection{C.2}{Verified Factor Groups}{app:verified_factor_groups}
\appTOCsubsection{C.3}{Support-Verification Results}{app:support_verification_results}
\appTOCsection{D.}{Statistical Test 2: Full-Dimensional Joint Distribution Support}{app:joint_distribution_support}
\appTOCsection{E.}{Interventional Action Set}{app:sensitivity}
\appTOCsection{F.}{Model Explanation}{app:model_details}
\appTOCsubsection{F.1}{End-to-End World Models}{app:end-to-end-world-models}
\appTOCsubsection{F.2}{Predictor-Only Models from True States}{app:predictor_only_models}
\appTOCsection{G.}{Training Details}{app:training-details}
\appTOCsection{H.}{Detailed Planning and Evaluation Configuration}{app:evaluation}
\appTOCsubsection{H.1}{Planning Evaluation Criteria}{app:planning-success-criteria}
\appTOCsubsection{H.2}{Configuration}{app:evaluation-configuration}
\appTOCsection{I.}{Results}{app:results}
\appTOCsubsection{I.1}{Full Results}{app:full-results}
\appTOCsubsection{I.2}{Factor Support versus Observation Support}{app:factor_vs_obs_support}
\appTOCsubsection{I.3}{Additional Observations}{app:additional-observations}
\appTOCsection{J.}{Toward Compositional Generalization in World Models}{app:future_compgen_theory}
\appTOCsubsection{J.1}{Identifying Reusable Mechanisms on the Training Support}{app:identifying_r}
\appTOCsubsection{J.2}{Extrapolating Identified Mechanisms to Novel Compositions}{app:extrapolating_i}
\appTOCsection{K.}{AnyGoal Library and Reproducibility}{app:anygoal_library}
\appTOCsubsection{K.1}{Installation and Dataset Access}{app:installation_a}
\appTOCsubsection{K.2}{World-Model Evaluation}{app:world-model_e}
\appTOCsubsection{K.3}{Reproducibility and Extensibility}{app:reproducibility}

\endgroup
\clearpage

\section{Compositional Generalization: Essential World-Model Test}
\label{app:compositional-world-model-test}

We formalize why our compositional evaluation is an essential test for world models. The key distinction is between extrapolating to unseen factors and recombining observed ones. Good action-conditioned world model should predict the
effect of an observed action on familiar physical factors, even when that particular combination was not selected by the behavior policy.

For clarity, consider a deterministic oracle-state setting with true controlled
dynamics $z_{t+1}=T^\star(z_t,a_t).$
Suppose the offline dataset is collected by a behavior policy \(a=\pi_E(z)\).
Then the observed state--action support lies on the policy graph,
\begin{equation}
\Gamma_{\pi_E}
=
\{(z,a): a=\pi_E(z)\}.
\label{eq:expert-graph}
\end{equation}
Thus, expert rollouts cover only the state--action pairings induced by the
behavior policy, not arbitrary interventions with observed actions.

\paragraph{Zero Training Error Does Not Imply Action-Conditioned Dynamics.}

A predictive world model trained on \((z,a,z')\sim p_{\mathrm{tr}}\) is
constrained only on the state--action pairs generated by the behavior policy.
Since observed actions satisfy \(a=\pi_E(z)\), the training loss constrains
\(\hat T(z,\pi_E(z))\), but not \(\hat T(z,a)\) for alternative observed actions
at the same state. Consequently, the action-ignorant predictor
\begin{equation}
\hat T_{\mathrm{deg}}(z,a)
=
T^\star(z,\pi_E(z))
\label{eq:degenerate-predictor}
\end{equation}
achieves zero prediction error on the behavior dataset while ignoring the
queried action \(a\).
Thus, expert-rollout prediction cannot distinguish a controlled dynamics model
from a predictor that only models behavior-policy transitions.

\paragraph{Compositional Evaluation Requires Action-Conditioned Dynamics}

Compositional evaluation rules out this degenerate solution. Assume actions have
nonzero effect within the observed marginal supports: for a set of oracle states
with positive measure, there exist observed actions \(a,a'\in\mathcal A^{\mathrm{tr}}\)
such that
\[
T^\star(z,a)\neq T^\star(z,a').
\]
Now consider any action-ignorant predictor \(\hat T(z,a)=h(z)\). If it were
correct on the relevant compositional state--action space, then for every
observed action \(a\in\mathcal A^{\mathrm{tr}}\), $h(z)=T^\star(z,a).$
Therefore, for any two observed actions \(a,a'\),
\begin{equation}
T^\star(z,a)
=
h(z)
=
T^\star(z,a'),
\label{eq:action-ignorant-contradiction}
\end{equation}
which contradicts the action-relevance assumption. Hence, no action-ignorant
model can be correct on recombinations of observed state and action factors when
actions causally affect the next state.

\begin{tcolorbox}[notebox]
\textbf{Interpretation.}
Expert rollouts test whether a model predicts what happens under the behavior
policy. Compositional state--action evaluation instead tests whether the model
predicts what happens when observed actions are applied to familiar physical
factors, including combinations not generated by the behavior policy. This is an
essential requirement for model-based control, since planning must evaluate
candidate actions other than the behavior action.
\end{tcolorbox}

\section{AnyGoal Task Details}
\label{app:data_detail}

\subsection{Two Room and Its Variants}
\label{app:two-room-variants}

\textbf{TwoRoom} is a simple navigation environment with two rooms connected by a narrow slot. The agent must move through the slot and reach a target location, with little to no meaningful interaction with the environment beyond collision with walls. Because the dynamics are essentially planar motion, this environment isolates whether models can preserve action-conditioned movement under goal-biased trajectory data.

\textbf{TwoRoom Route} is an interaction-light navigation variant. The agent must follow a piecewise-linear triangular route defined by a start point, an intermediate waypoint, and an invisible endpoint, rather than moving directly through the slot toward the target. Episodes mix clean, noisy, and variable-speed actions within the observed action support. This variant tests whether the model preserves planar action-conditioned dynamics when familiar positions and actions are recombined outside the dominant slot-crossing behavior of the base dataset.

\subsection{PushT and Its Variants}
\label{app:pusht-variants}

\textbf{PushT} is a planar manipulation environment where the agent must push a T-shaped block into a fixed T-shaped goal pose. Unlike TwoRoom, the base task already requires meaningful contact, object translation, and object rotation. The expert trajectories are strongly goal-directed, coupling agent motion with the block pose and the target configuration. This environment tests whether world models learn action-conditioned contact dynamics rather than only the dominant expert pushing manifold.

\textbf{AvoidT} is a non-interactive variant. The agent must avoid touching the physical T-block and move to one of four corner regions. This variant keeps the agent and action factors familiar, but changes the required behavior from goal-directed pushing to free-space navigation around the object. It primarily tests whether the model preserves the action effect when the expert pushing objective is no longer appropriate.

\textbf{StackT} is an interaction-rich variant. The physical T-block must be placed so that its head overlaps with the fixed goal T, while its rotation is opposite to the original PushT goal orientation, creating a plus-shaped target configuration. This variant requires contact and rotation dynamics under a goal that differs from the original training objective.

\textbf{PushT Anti-goal} is also interaction-rich and directly reverses the dominant goal bias in the base dataset. The physical T-block is initialized near the original PushT goal pose, but the agent must push it away from the fixed goal region and move it toward one of four corner regions, with rotation opposite to the original goal orientation. This tests whether the model can predict familiar pushing dynamics when the required behavior contradicts the original expert trajectory distribution.

\subsection{Insert-Peg and Its Variants}
\label{app:insert-peg-variants}

\textbf{Insert-Peg} is a planar manipulation environment where an arm must move a peg toward a sampled slot and insert it into the target opening. The task involves articulated robot motion, object contact, and alignment between the peg and the slot. Expert trajectories are strongly structured around successful insertion behavior. This environment tests whether models can learn action-conditioned manipulation dynamics under goal-biased contact-rich data.

\textbf{Avoid-Peg} is an interaction-light variant. The arm must move around the target box in a specified hugging direction while avoiding contact with the peg. This changes the goal-conditioned arm behavior without requiring peg manipulation, testing whether the model can predict familiar arm motion under a non-insertion objective.

\textbf{Extract-Peg} is an interaction-rich variant. The arm must extract the peg from the target box, move it through a waypoint, and place it at a destination, with episodes balanced between left- and right-side box placements. This reverses the dominant insertion behavior while keeping the peg, box, workspace, and action space familiar.

\subsection{OGBench Cube and Its Variants}
\label{app:ogbench-cube-variants}

\textbf{Cube} is a robot manipulation environment where the end-effector must interact with a cube and move it toward a target configuration. The base behavior follows a structured pick-and-place pattern, including approach, grasp, lift, transport, and release. This creates a goal-biased trajectory distribution in which robot motion, gripper state, cube pose, and target location are tightly coupled. The environment tests whether models learn manipulation dynamics beyond this dominant expert sequence.

\textbf{CubeHop} is a non-interactive variant. The robot does not intentionally move the cube. Instead, the end-effector moves between waypoint and target positions while gradually changing its yaw. This variant changes the goal-conditioned end-effector behavior while avoiding object interaction. It tests whether the model can predict familiar robot motion and yaw changes outside the dominant cube-manipulation trajectory distribution.

\textbf{CubeDiagonal} is an interaction-rich variant. The robot manipulates the cube using a diagonal grasp configuration, where the gripper contacts opposite edges of the cube instead of the standard parallel faces. This introduces a novel contact regime with different grasp geometry while keeping the object, task objective, and action space unchanged. It tests whether the model can compose familiar manipulation factors under unseen interaction configurations.

\section{Statistical Test 1: Marginal Distribution Support}
\label{app:distribution_support_criteria}

\subsection{Distribution Support Criteria}
\label{app:distribution-support-criteria-details}
We describe the statistical criteria used to decide whether a query
distribution remains inside the support of a reference distribution. Let
$P=\{p_n\}_{n=1}^N$ denote the reference or training point set, and let
$Q=\{q_m\}_{m=1}^M$ denote the query point set.

For each factor, we compute the empirical mean
$\mu_P=\frac{1}{N}\sum_{n=1}^N p_n$ and regularized covariance
$\Sigma_P=\operatorname{Cov}(P)+\lambda I$, where $\lambda>0$ is a small
diagonal regularizer used to ensure that the covariance matrix is invertible.
This is particularly important for low-rank or manifold-valued factors, such as
circular variables represented by sine and cosine coordinates.

Now we measure three metric respectively.
First, the Mahalanobis criterion checks whether query points remain within the reference distribution's main covariance structure.

\begin{tcolorbox}[notebox]
\textbf{Empirical Mahalanobis criterion.}
The empirical squared Mahalanobis distance of a point $x$ from the reference
distribution is
\begin{equation}
d_P^2(x)
=
(x-\mu_P)^\top \Sigma_P^{-1}(x-\mu_P).
\end{equation}
We compute the reference threshold as the $99$th percentile of the empirical
Mahalanobis distances of the reference points,
$\tau_{0.99}=\operatorname{Quantile}_{0.99}
(\{d_P^2(p_n)\}_{n=1}^N)$.
The query set passes the Mahalanobis support criterion if at least $95\%$ of
its points lie below this reference threshold:
\begin{equation}
\operatorname{Mahalanobis}_{0.99}(Q;P)
=
\left[
\frac{1}{M}\sum_{m=1}^M
\mathbf{1}\!\left[d_P^2(q_m)\le \tau_{0.99}\right]
\ge 0.95
\right].
\end{equation}
For a one-dimensional factor, this criterion reduces to the regularized squared
$z$-score $d_P^2(x)=(x-\mu_P)^2/(\sigma_P^2+\lambda)$.
\end{tcolorbox}

Second, the KDE criterion checks whether query points lie in regions with sufficient empirical reference density, beyond an ellipsoidal approximation.

\begin{tcolorbox}[notebox]
\textbf{Histogram-binned KDE criterion.}
We estimate a reference density \(\rho_P\) with a histogram-binned Gaussian KDE
fit only on the reference set \(P\). Query points are then evaluated under this
fixed density estimate. The histogram grid is constructed from the pooled
reference--query range with a small padding, and Gaussian smoothing uses
Scott's rule with a lower bandwidth clip to avoid degenerate bins.

After fitting the KDE, we evaluate the density of all reference samples and set
the threshold to the empirical first percentile,
\[
\delta_{0.01}
=
\operatorname{Quantile}_{0.01}
\left(\{\rho_P(p_n)\}_{n=1}^N\right).
\]
The query set passes the KDE criterion if at least \(95\%\) of query points have
density no smaller than this reference-sample threshold:
\begin{equation}
\operatorname{KDE}_{0.01}(Q;P)
=
\left[
\frac{1}{M}\sum_{m=1}^M
\mathbf{1}\!\left[\rho_P(q_m)\ge \delta_{0.01}\right]
\ge 0.95
\right].
\end{equation}
For periodic yaw variables, KDE smoothing is applied with wrapped boundary
conditions.
\end{tcolorbox}

Lastly, the range criterion checks whether query points avoid direct coordinate-wise extrapolation beyond the observed marginal range.
\begin{tcolorbox}[notebox]
\textbf{Observed-range criterion.}
As a complementary support check, we compute an axis-wise observed-range test.
For each query point $q_m$, we check whether all of its coordinates lie inside
the coordinate-wise minimum and maximum of the reference set, i.e.,
$q_m\in[\min(P),\max(P)]$ iff
$\min_n p_{n,j}\le q_{m,j}\le \max_n p_{n,j}$ for all $j$.
The observed-range criterion is
\begin{equation}
\operatorname{range}(Q;P)
=
\left[
\frac{1}{M}\sum_{m=1}^M
\mathbf{1}\!\left[q_m\in[\min(P),\max(P)]\right]
\ge 0.99
\right].
\end{equation}
For circular variables, such as yaw, the observed-range criterion is ignored
because the interval boundary is representation-dependent. In that case, the
Mahalanobis and KDE criteria are used without the range check.
\end{tcolorbox}

The final verdict is the conjunction of all non-ignored criteria:
\begin{equation}
\operatorname{pass}(Q;P)
=
\operatorname{range}(Q;P)
\wedge
\operatorname{Mahalanobis}_{0.99}(Q;P)
\wedge
\operatorname{KDE}_{0.01}(Q;P).
\end{equation}
If the observed-range criterion is ignored for a circular factor, the verdict is
computed over the remaining criteria:
\begin{equation}
\operatorname{pass}(Q;P)
=
\operatorname{Mahalanobis}_{0.99}(Q;P)
\wedge
\operatorname{KDE}_{0.01}(Q;P).
\end{equation}

This design separates marginal factor support from full trajectory-distribution
typicality. The support checks are applied at the level of semantic factor
groups to verify that each physical factor remains within the reference support.
Combinatorial novelty can still arise when individually supported factors
appear in new joint configurations that are rare or absent under the expert
trajectory distribution.

\subsection{Verified Factor Groups}
\label{app:verified_factor_groups}

We apply the three support criteria above to task-relevant factor groups rather than to the full joint state--action vector. This verifies marginal coverage of the physical factors and actions used in each goal-shifted variant, while allowing the full combinations to remain novel. Table~\ref{tab:support_verification_groups} lists the factor groups checked for each reference--query pair.

\begin{table}[h]
\centering
\small
\setlength{\tabcolsep}{3pt}
\renewcommand{\arraystretch}{1.15}
\caption{\textbf{Every independent factor is in-distribution with respect to the reference dataset.} Factor groups used for support verification. For each reference dataset
\(P\), we compare the corresponding goal-shifted variant datasets \(Q\) against
the listed task-relevant factors..}
\label{tab:support_verification_groups}
\begin{tabularx}{\linewidth}{
>{\raggedright\arraybackslash}p{0.20\linewidth}
>{\raggedright\arraybackslash}p{0.25\linewidth}
>{\raggedright\arraybackslash}X}
\toprule
Reference dataset \(P\) & Query dataset \(Q\) & Factor groups checked \\
\midrule

TwoRoom
& TR-Route
& Agent position \((2\mathrm{D})\); action position delta \((2\mathrm{D})\) \\

PushT
& Avoid-T; Stack-T; Push-T Antigoal
& Agent position \((2\mathrm{D})\); block position \((2\mathrm{D})\);
  action \((2\mathrm{D})\); block angle. \\

Insert-Peg
& Avoid-Peg; Extract-Peg
& Arm root angle;
Arm shoulder angle;
Arm elbow angle;
Arm wrist angle;
Finger angle;
Fingertip angle;
Thumb angle;
Thumbtip angle;
hand, peg, and target position (2D); hand, peg, and target orientation (2D) ;
5 motor commands ($1\mathrm{D}$) independently. \\

Cube
& Cube-Hop; Cube-Diagonal
& End-effector position \((3\mathrm{D})\); block position \((3\mathrm{D})\);
  block quaternion \((4\mathrm{D})\); action position delta \((3\mathrm{D})\);
  gripper opening; gripper contact; end-effector yaw; action yaw delta;
  action gripper delta. ; 6 joint pos independently\\

\bottomrule
\end{tabularx}
\end{table}

\subsection{Support-Verification Results}
\label{app:support_verification_results}

Table~\ref{tab:distribution_support_full} reports the full support-verification results for the factor groups in Table~\ref{tab:support_verification_groups}. For each reference--query pair, we report the pass rate under the Mahalanobis, KDE, and observed-range criteria, together with the final conjunction verdict. All checked factors pass, confirming that the AnyGoal variants remain marginally in distribution with respect to the corresponding base datasets.

{\small
\setlength{\tabcolsep}{3pt}
\renewcommand{\arraystretch}{0.9}

\begin{longtable}[h]{@{}llp{1.55in}cccc@{}}
\caption{Distribution-support verification between each reference dataset ($P$) and query dataset ($Q$). A dash indicates that the range criterion is not applied to circular variables.(\textbf{All \textcolor{ForestGreen}{Passed}}) }
\label{tab:distribution_support_full}\\
\toprule
$P$ & $Q$ & Factor of variants & Mahalanobis 0.99 & KDE & Range & Final verdict \\
\midrule
\endfirsthead

\multicolumn{7}{@{}l}{\small\itshape Table~\ref{tab:distribution_support_full} continued.}\\
\toprule
$P$ & $Q$ & Factor of variants & Mahalanobis 0.99 & KDE & Range & Final verdict \\
\midrule
\endhead

\midrule
\multicolumn{7}{r@{}}{\small\itshape Continued on next page}\\
\endfoot

\bottomrule
\endlastfoot

\multirow{2}{*}{TwoRoom} & \multirow{2}{*}{TR-Route} & agent pos (2D) & 0.970 & 1.000 & 1.000 & pass \\
 &  & action (2D) & 1.000 & 0.957 & 1.000 & pass \\
\midrule
\multirow{4}{*}{PushT} & \multirow{4}{*}{Avoid-T} & block angle & 0.970 & 0.963 & -- & pass \\
 &  & agent pos (2D) & 0.968 & 0.984 & 1.000 & pass \\
 &  & block pos (2D) & 1.000 & 0.990 & 1.000 & pass \\
 &  & action (2D) & 1.000 & 0.995 & 1.000 & pass \\
\midrule
\multirow{4}{*}{PushT} & \multirow{4}{*}{Stack-T} & block angle & 0.998 & 0.989 & -- & pass \\
 &  & agent pos (2D) & 0.992 & 0.989 & 1.000 & pass \\
 &  & block pos (2D) & 1.000 & 0.999 & 1.000 & pass \\
 &  & action (2D) & 1.000 & 0.954 & 1.000 & pass \\
\midrule
\multirow{4}{*}{PushT} & \multirow{4}{*}{Push-T Antigoal} & block angle & 1.000 & 0.988 & -- & pass \\
 &  & agent pos (2D) & 0.988 & 0.986 & 1.000 & pass \\
 &  & block pos (2D) & 1.000 & 1.000 & 1.000 & pass \\
 &  & action (2D) & 1.000 & 0.989 & 1.000 & pass \\
\midrule
\multirow{19}{*}{Insert-Peg} & \multirow{19}{*}{Avoid-Peg} & arm root angle & 0.980 & 0.963 & 1.000 & pass \\
 &  & arm shoulder angle & 1.000 & 0.999 & 1.000 & pass \\
 &  & arm elbow angle & 0.999 & 0.998 & 1.000 & pass \\
 &  & arm wrist angle & 1.000 & 1.000 & 1.000 & pass \\
 &  & finger angle & 1.000 & 1.000 & 1.000 & pass \\
 &  & fingertip angle & 0.996 & 0.998 & 1.000 & pass \\
 &  & thumb angle & 1.000 & 1.000 & 1.000 & pass \\
 &  & thumbtip angle & 0.994 & 0.997 & 1.000 & pass \\
 &  & hand position (x,z) & 0.978 & 1.000 & 1.000 & pass \\
 &  & hand orientation (qw,qy) & 1.000 & 1.000 & 1.000 & pass \\
 &  & peg position (x,z) & 1.000 & 1.000 & 1.000 & pass \\
 &  & peg orientation (qw,qy) & 0.999 & 0.999 & 1.000 & pass \\
 &  & target position (x,z) & 1.000 & 1.000 & 1.000 & pass \\
 &  & target orientation (qw,qy) & 1.000 & 1.000 & 1.000 & pass \\
 &  & motor command 0 & 1.000 & 0.982 & 1.000 & pass \\
 &  & motor command 1 & 1.000 & 1.000 & 1.000 & pass \\
 &  & motor command 2 & 1.000 & 0.991 & 1.000 & pass \\
 &  & motor command 3 & 1.000 & 1.000 & 1.000 & pass \\
 &  & motor command 4 & 1.000 & 1.000 & 1.000 & pass \\
\midrule
\multirow{19}{*}{Insert-Peg} & \multirow{19}{*}{Extract-Peg} & arm root angle & 0.996 & 0.975 & 1.000 & pass \\
 &  & arm shoulder angle & 0.989 & 0.983 & 1.000 & pass \\
 &  & arm elbow angle & 0.988 & 0.988 & 1.000 & pass \\
 &  & arm wrist angle & 1.000 & 0.998 & 1.000 & pass \\
 &  & finger angle & 1.000 & 1.000 & 1.000 & pass \\
 &  & fingertip angle & 0.989 & 0.991 & 1.000 & pass \\
 &  & thumb angle & 1.000 & 0.997 & 1.000 & pass \\
 &  & thumbtip angle & 0.966 & 0.960 & 1.000 & pass \\
 &  & hand position (x,z) & 1.000 & 0.998 & 1.000 & pass \\
 &  & hand orientation (qw,qy) & 1.000 & 1.000 & 1.000 & pass \\
 &  & peg position (x,z) & 0.998 & 0.997 & 1.000 & pass \\
 &  & peg orientation (qw,qy) & 0.997 & 0.997 & 1.000 & pass \\
 &  & target position (x,z) & 0.983 & 0.978 & 1.000 & pass \\
 &  & target orientation (qw,qy) & 1.000 & 1.000 & 1.000 & pass \\
 &  & motor command 0 & 1.000 & 1.000 & 1.000 & pass \\
 &  & motor command 1 & 1.000 & 1.000 & 1.000 & pass \\
 &  & motor command 2 & 1.000 & 0.987 & 1.000 & pass \\
 &  & motor command 3 & 1.000 & 0.998 & 1.000 & pass \\
 &  & motor command 4 & 1.000 & 1.000 & 1.000 & pass \\
 \midrule
\multirow{15}{*}{Cube} & \multirow{15}{*}{Cube-Hop} & joint pos 0 & 0.995 & 0.995 & 1.000 & pass \\
 &  & joint pos 1 & 0.997 & 0.998 & 1.000 & pass \\
 &  & joint pos 2 & 0.974 & 0.974 & 1.000 & pass \\
 &  & joint pos 3 & 0.990 & 0.991 & 1.000 & pass \\
 &  & joint pos 4 & 1.000 & 1.000 & 1.000 & pass \\
 &  & joint pos 5 & 0.957 & 0.960 & 1.000 & pass \\
 &  & gripper opening & 0.980 & 1.000 & 1.000 & pass \\
 &  & gripper contact & 1.000 & 1.000 & 1.000 & pass \\
 &  & EE yaw & 0.958 & 0.961 & -- & pass \\
 &  & delta yaw & 1.000 & 1.000 & 1.000 & pass \\
 &  & delta gripper & 1.000 & 1.000 & 1.000 & pass \\
 &  & EE pos (3D) & 0.969 & 0.970 & 1.000 & pass \\
 &  & block pos (3D) & 1.000 & 1.000 & 1.000 & pass \\
 &  & block quat (4D) & 1.000 & 1.000 & 1.000 & pass \\
 &  & delta pos (3D) & 1.000 & 1.000 & 1.000 & pass \\
\midrule
\multirow{15}{*}{Cube} & \multirow{15}{*}{Cube-Diagonal} & joint pos 0 & 0.989 & 0.990 & 1.000 & pass \\
 &  & joint pos 1 & 0.992 & 0.991 & 1.000 & pass \\
 &  & joint pos 2 & 0.990 & 0.990 & 1.000 & pass \\
 &  & joint pos 3 & 0.991 & 0.992 & 1.000 & pass \\
 &  & joint pos 4 & 0.978 & 0.979 & 1.000 & pass \\
 &  & joint pos 5 & 1.000 & 1.000 & 1.000 & pass \\
 &  & gripper opening & 0.989 & 0.984 & 1.000 & pass \\
 &  & gripper contact & 1.000 & 0.985 & 1.000 & pass \\
 &  & EE yaw & 1.000 & 1.000 & -- & pass \\
 &  & delta yaw & 1.000 & 0.992 & 1.000 & pass \\
 &  & delta gripper & 0.995 & 0.995 & 1.000 & pass \\
 &  & EE pos (3D) & 0.989 & 0.990 & 1.000 & pass \\
 &  & block pos (3D) & 0.991 & 0.990 & 1.000 & pass \\
 &  & block quat (4D) & 0.997 & 0.987 & 1.000 & pass \\
 &  & delta pos (3D) & 0.989 & 0.990 & 1.000 & pass \\
\end{longtable}
}

\section{Statistical Test 2: Full-Dimensional Joint Distribution Support}
\label{app:joint_distribution_support}

We complement the factor-wise support tests in
Table~\ref{tab:distribution_support_full} with a full-dimensional joint test.
The factor-wise tests verify whether each task-relevant factor or factor group
is marginally supported by the reference data. In contrast, the joint test asks
whether the complete state--action combination lies in a high-density region of
the reference distribution. This test uses the concatenated vectors directly,
without coordinate marginalization or worst-pair reduction.

For each transition, let \(a\) denote the action, \(s_{\mathrm{agent}}\) the
agent-observed state, and \(s_{\mathrm{other}}\) the task-object and target
state. We evaluate three joint vectors:
\[
z_{\mathrm{agent}} = s_{\mathrm{agent}}, \qquad
z_{\mathrm{all}} = [s_{\mathrm{agent}}, s_{\mathrm{other}}], \qquad
z_{a,\mathrm{all}} = [a, s_{\mathrm{agent}}, s_{\mathrm{other}}].
\]
Only rows that are finite in every coordinate of the corresponding vector are
retained. Thus, terminal rows with undefined actions contribute to the
state-only vectors but not to the action--state vector.

\begin{table}[h]
\centering
\small
\caption{Components and dimensionalities of the full-dimensional joint KDE.}
\label{tab:joint-vector-definitions}
\begin{tabular}{llll}
\toprule
Family & \(s_{\mathrm{agent}}\) & \(s_{\mathrm{all}}\) & \([a,s_{\mathrm{all}}]\) \\
\midrule
TwoRoom & position (2D) & position (2D) & 4D \\
PushT & agent position (2D) & \(+\) block pose (5D total) & 7D \\
Cube & EE/gripper state (6D) & \(+\) block pose (13D total) & 18D \\
Manipulator & joints \(+\) hand pose (20D) & \(+\) peg/target poses (28D total) & 33D \\
\bottomrule
\end{tabular}
\end{table}

For each reference vector set, we quantize each coordinate using reference-based
limits with small padding, and assign zero density to target values outside
these limits. We then whiten the quantized vectors using the full reference
covariance with a small diagonal regularizer. This preserves correlations
between action and state coordinates when computing kernel distances.

We fit a full-dimensional Gaussian KDE to the whitened reference vectors using
Scott's bandwidth rule. The density is evaluated without reference subsampling,
dimension dropping, nearest-neighbor truncation, or pairwise marginalization.
The density threshold is calibrated only from the reference samples:
\begin{equation}
\tau
=
Q_{0.01}\!\left(
\{\log \widehat p(x_i)\}_{i=1}^{N}
\right).
\end{equation}
For target vectors \(Y=\{y_j\}_{j=1}^{M}\), we report the fraction whose
log-density exceeds this reference-sample threshold:
\begin{equation}
\operatorname{ID}(Y;X)
=
\frac{1}{M}
\sum_{j=1}^{M}
\mathbf 1\!\left[
\log \widehat p(y_j) \geq \tau
\right].
\end{equation}

The joint score is therefore a full-vector KDE support percentage. Individual
factors can pass their marginal support tests while the concatenated vector has
a low joint score, indicating that the corresponding combination of familiar
values is rare under the reference distribution.

\section{Interventional Action Set}
\label{app:sensitivity}

For each environment, we choose a fixed interventional action set that remains within the empirical action support while covering the task-relevant intervention directions as uniformly as possible. For environments with direct 2D controls, this is a regular grid in action space. For higher-dimensional manipulation actions, we vary only the dimensions needed for the intended spatial intervention and keep the remaining dimensions fixed.

\paragraph{TwoRoom.}
For TwoRoom, actions are two-dimensional position deltas, so interventional
interventions are defined directly in the action space. We use a fixed
\(3\times3\) grid that is uniformly spread over the two action dimensions and
includes the zero action as the center intervention.

\begin{tcolorbox}[notebox]
\textbf{Selected TwoRoom interventional actions.}
\[
\mathcal A_{\mathrm{cf}}^{\mathrm{TwoRoom}}
=
\left\{
\begin{array}{rr}
-1 & -1 \\
-1 &  0 \\
-1 &  1 \\
 0 & -1 \\
 0 &  0 \\
 0 &  1 \\
 1 & -1 \\
 1 &  0 \\
 1 &  1
\end{array}
\right\}
=
\{-1,0,1\}\times\{-1,0,1\}.
\]
Each row is a 2D action.
\end{tcolorbox}

\paragraph{Push-T.}
For Push-T, actions are also two-dimensional planar controls. We therefore use
a fixed \(3\times3\) grid in the planar action space. The grid is chosen to
provide evenly distributed in-distribution interventions at the empirical action
scale of the dataset.

\begin{tcolorbox}[notebox]
\textbf{Selected Push-T interventional actions.}
\[
\mathcal A_{\mathrm{cf}}^{\mathrm{PushT}}
=
\left\{
\begin{array}{rr}
 0    &  0    \\
 0.25 &  0    \\
-0.25 &  0    \\
 0.25 & -0.25 \\
 0.25 &  0.25 \\
-0.25 & -0.25 \\
-0.25 &  0.25 \\
 0    & -0.25 \\
 0    &  0.25
\end{array}
\right\}.
\]
Each row is a 2D planar action.
\end{tcolorbox}

\paragraph{Insert-Peg.}
Unlike the 2D navigation and planar pushing tasks, Insert-Peg actions are
5-dimensional motor commands rather than direct task-space displacements.
Therefore, interventional actions cannot be chosen by simply placing actions on
a Euclidean grid in task space. Instead, we select actions by their realized
effect on the hand endpoint.

We first choose a representative state near the center of the expert-state
distribution. Each candidate action is sampled from the dataset and repeated
for five environment control steps from this same state. We then measure the
resulting hand endpoint in the \(x\)-\(z\) plane. To ensure that interventions
remain in distribution, we retain only actions whose 20-nearest-neighbor
distance in the raw 5D action space is below the 95th-percentile threshold
estimated from held-out dataset actions.

Among the retained actions, we select a fixed set of nine interventional
actions whose realized endpoints form the widest approximately uniform
\(3\times3\) grid. The zero action is assigned to the center, and the remaining
actions are chosen to maximize grid spacing while keeping each endpoint within
\(10\%\) of its corresponding ideal grid point. The resulting grid spacing is
\(0.0358\), with a maximum endpoint mismatch of \(9.95\%\) of the grid spacing.

\begin{tcolorbox}[notebox]
\textbf{Selected Insert-Peg actions.}
\[
\mathcal A_{\mathrm{cf}}^{\mathrm{Insert}}
=
\left\{
\begin{array}{rrrrr}
 0.6728 & -0.9814 & -0.9932 & -1.0000 &  0.9900 \\
 0.1353 & -1.0000 &  0.6557 & -0.9972 &  1.0000 \\
-0.6023 & -0.1204 &  0.3142 &  0.9943 & -1.0000 \\
 1.0000 & -0.8002 & -1.0000 & -1.0000 &  0.9739 \\
 0.0000 &  0.0000 &  0.0000 &  0.0000 &  0.0000 \\
-0.7797 &  1.0000 & -0.6672 &  0.6553 & -1.0000 \\
 0.9130 & -0.0082 & -1.0000 & -0.9820 &  1.0000 \\
-0.0513 &  1.0000 & -1.0000 &  0.5704 &  0.1936 \\
-0.5833 &  0.8958 &  0.8234 &  0.0582 &  0.1618
\end{array}
\right\}.
\]
Each row is a 5D motor command. The zero action is the center intervention.
\end{tcolorbox}

\paragraph{Cube.}
For Cube, the raw action is five-dimensional, but the interventional protocols
are designed to probe spatial sensitivity of end-effector motion. We therefore
vary only translational action dimensions and keep \(\Delta\mathrm{yaw}\) and
\(\Delta\mathrm{gripper}\) fixed at zero. The horizontal protocol uses an
\(x\)-\(y\) grid.

\begin{tcolorbox}[notebox]
\textbf{Selected Cube horizontal interventional actions.}
\[
\mathcal A_{\mathrm{cf}}^{\mathrm{Cube\text{-}xy}}
=
\left\{
\begin{array}{rrrrr}
-0.5427 & -0.9504 & 0 & 0 & 0 \\
-0.5427 &  0      & 0 & 0 & 0 \\
-0.5427 &  0.9504 & 0 & 0 & 0 \\
 0      & -0.9504 & 0 & 0 & 0 \\
 0      &  0      & 0 & 0 & 0 \\
 0      &  0.9504 & 0 & 0 & 0 \\
 0.5427 & -0.9504 & 0 & 0 & 0 \\
 0.5427 &  0      & 0 & 0 & 0 \\
 0.5427 &  0.9504 & 0 & 0 & 0
\end{array}
\right\}.
\]
Each row is a normalized 5D Cube action
\((a_x,a_y,a_z,\Delta\mathrm{yaw},\Delta\mathrm{gripper})\).
\end{tcolorbox}


\paragraph{Implementation Detail.}
\label{app:sensitivity-implementation-detail}

Within orignal LeWM architecture, we measure
action sensitivity using hidden representations from the predictor blocks. Let
\(f_l\) denote the last-token hidden representation after the \(l\)-th
ConditionalBlock,
\[
f_l = \texttt{output[:, -1, :]}
\]
obtained from a forward hook on
\(\texttt{model.predictor.transformer.layers[layer\_idx]}\).
Thus, \(f_l\) is not the representation immediately after AdaLN modulation; it
is the output of the full AdaLN-conditioned transformer block, including the
attention and MLP residual updates. In all reported layer-wise analyses, we
average the sensitivity scores across these block outputs. For the final
prediction representation, we use \(\texttt{pred[:, -1, :]}\), the output of
\(\texttt{model.predict(...)}\) after the final \(\texttt{pred\_proj}\) layer.

\section{Model Explanation}
\label{app:model_details}

Our model selection is designed to span several substantially different
approaches to visual world modeling rather than to concentrate on a single
architectural family. In particular, we evaluate both joint-embedding
predictive architectures (JEPAs) and generative visual world models. Within the
JEPA family, we include methods that jointly learn perception and dynamics from
the offline dataset as well as a method that keeps a large pretrained visual
encoder frozen. Within the generative family, we include both a direct
action-conditioned generator and a latent video diffusion model with an explicit
visual encoder--decoder. This diversity is important for our central question:
whether goal-induced combinatorial bias is specific to a particular
representation or generative objective, or can arise more broadly when
action-conditioned prediction is trained on a biased offline distribution. For consistency, all models use frame skip \(5\), history length \(3\), and one-step prediction, except Sensorimotor-WM, where we keep the original history length \(1\) to preserve its inverse-dynamics setup.
\subsection{End-to-End World Models}
\label{app:end-to-end-world-models}

\paragraph{LeWorldModel (LeWM).}
LeWorldModel~\citep{lewm} is an end-to-end joint-embedding predictive
architecture trained directly from pixels and actions. An image encoder maps
observations to compact latent representations, and an autoregressive
transformer predictor models the next latent representation conditioned on the
action. Actions are injected into each transformer layer through adaptive layer
normalization (AdaLN-zero). Unlike approaches that rely on a frozen encoder,
stop-gradient target network, or exponential-moving-average encoder, LeWM
jointly optimizes its encoder and predictor.

A central design choice in LeWM is its anti-collapse regularization. In addition
to the next-embedding mean-squared prediction loss, LeWM applies the
Sketched-Isotropic-Gaussian Regularizer (SIGReg), which encourages the empirical
latent distribution to match an isotropic Gaussian. The resulting model is
particularly useful for our study because it represents a recent example of a
fully end-to-end JEPA in which both the visual representation and dynamics
predictor are shaped by the same offline trajectories.

\paragraph{Sensorimotor World Model (SMWM).}
The Sensorimotor World Model~\citep{sensori-wm} is also an end-to-end JEPA, but uses a
different principle to prevent representation collapse. Rather than prescribing
the marginal geometry of the latent representation, SMWM introduces an inverse
dynamics model that predicts the executed action from consecutive latent states.
The encoder, forward dynamics model, and inverse model are optimized jointly
using forward-prediction and inverse-dynamics losses.

The inverse objective requires consecutive representations to preserve
information that is useful for recovering the action that produced the
transition. Consequently, the learned representation is explicitly biased
toward controllable, action-relevant structure. Architecturally, SMWM adopts the
LeWM encoder and transformer forward predictor, including AdaLN-zero action
conditioning, while replacing Gaussian latent regularization with inverse
dynamics as its anti-collapse mechanism. This makes SMWM particularly relevant
to our benchmark: it provides a strong test of whether explicitly encouraging
action-aware representations is sufficient to prevent failures induced by
goal-biased state--action coverage.

\paragraph{DINO World Model (DINO-WM).}
DINO-WM~\citep{dinowm} separates representation learning from dynamics learning.
Instead of jointly learning an image encoder from the offline control dataset,
it uses pretrained DINOv2 patch features as the observation representation and
keeps the DINOv2 encoder frozen during both world-model training and evaluation.
A causal ViT-based transition model then predicts future patch representations
from a history of latent observations and actions.

DINO-WM also differs in how actions enter the predictor. The action is first
mapped through an MLP and the resulting action representation is concatenated to
each spatial patch representation before the transformer processes the latent
state. Thus, compared with LeWM and SMWM, DINO-WM provides a useful separation
between perception and dynamics: the visual representation is inherited from
large-scale pretraining, while only the action-conditioned transition model is
learned from the potentially goal-biased offline trajectories. This lets us test
whether combinatorial failures persist even when the visual encoder itself is
not learned from the biased task dataset.

\paragraph{DriftWorld.}
DriftWorld~\citep{driftworld} represents a substantially different,
reconstruction-based family of world models. It is an action-conditioned
generative model based on drifting rather than a JEPA or iterative diffusion
model. Its core generator is a spatial--temporal U-Net that receives observation
history, Gaussian noise, and a future action sequence and generates the
corresponding future visual observations. Actions condition the U-Net
frame-by-frame through FiLM, such that each future frame is explicitly
conditioned on its corresponding action.

For relatively simple simulated domains such as PushT, DriftWorld operates
directly in pixel space and does not require a learned visual state encoder
between the observations and the action-conditioned generator. Its drifting
objective moves generated samples toward ground-truth future observations and
away from generated negatives, enabling future video generation in a single
forward pass at inference time. DriftWorld therefore provides an important
contrast to the latent JEPA models: prediction and visual generation are
implemented by a single action-conditioned generative backbone rather than an
explicit encoder followed by a latent dynamics predictor.

\paragraph{IRASim.}
IRASim~\citep{irasim} represents the latent generative end of our model suite.
It formulates world modeling for robot manipulation as
trajectory-conditioned video generation. Frames are first compressed using a
pretrained VAE, after which a Diffusion Transformer (DiT) models future video
latents conditioned on historical observations and a robot action trajectory.
The generated latents are subsequently decoded back to RGB observations.

A defining component of IRASim is its fine-grained action conditioning. Rather
than conditioning the complete video only on a global trajectory embedding,
IRASim can associate individual actions with their corresponding generated
frames through frame-level adaptive normalization. This architecture gives us a
generative model that is structurally different from DriftWorld: visual
prediction occurs in an encoded latent space, the dynamics generator is a
diffusion transformer, and an explicit encoder--decoder surrounds the dynamics
model.

\subsection{Predictor-Only Models from True States}
\label{app:predictor_only_models}

End-to-end evaluation mixes perception and dynamics-prediction errors. To isolate the latter, we use predictor-only models that replace the visual encoder with oracle states from simulator metadata and proprioceptive signals. These oracle states contain task-relevant physical variables, such as positions, orientations, agent or gripper configurations, and other environment-specific state factors needed for one-step prediction.

We refer to these inputs as oracle state representations rather than unique or minimal Markov states. Their role is operational: they provide the predictor with relevant physical information directly, removing visual representation learning as the explanation for prediction failure. We instantiate three predictor families derived from the end-to-end models above. The two transformer-based predictor-only models are trained with next-state mean-squared error,
\begin{equation}
    \mathcal L_{\mathrm{pred}}=
\left\|
\hat s_{t+1}-s_{t+1}
\right\|_2^2.
\end{equation}
Toy-DriftWorld instead retains the drifting particle loss used by its generative backbone.

\paragraph{AdaLN Transformer Predictor.}
Our first predictor follows the forward-model architecture shared by LeWM and
SMWM. Ground-truth state factors replace the learned visual embedding, while the
action is injected into the transformer through AdaLN-zero. The transformer
predicts the next true state directly. Because LeWM and SMWM use the same
forward-predictor family, a single predictor-only architecture represents both
models in this controlled setting. Their end-to-end distinction---SIGReg versus
inverse-dynamics representation regularization---is intentionally removed once
the encoder is replaced by the oracle state.

\paragraph{Action-Duplicated Transformer Predictor.}
Our second predictor follows DINO-WM's transition-model conditioning strategy.
The true state replaces DINO patch features, while an embedded action is
attached to the state representation before processing by a causal transformer.
This preserves the central architectural distinction of DINO-WM's predictor:
action information enters through the predictor input rather than modulating
transformer layers through AdaLN. The model is again trained only to regress the
next true state with MSE.

\paragraph{DriftWorld-Style Generative Predictor.}
Our third predictor is derived from the action-conditioned generative backbone
of DriftWorld. Because DriftWorld does not require a separate visual encoder for
the simple simulated setting, its U-Net constitutes the principal predictive
computation. For the predictor-only experiment, we use a reduced-size version
of this backbone and replace pixel observations and outputs with the
corresponding true-state representation. The predictor retains the drifting particle loss while operating on oracle-state inputs and next-state targets.

\subsubsection{Oracle States}
\label{app:oracle-states}

Oracle-state predictors replace RGB observations with task-relevant physical state variables from each base environment.

\begin{table}[h]
\centering
\small
\setlength{\tabcolsep}{4pt}
\renewcommand{\arraystretch}{1.15}
\caption{Oracle state variables used by predictor-only models for each base environment.}
\label{tab:oracle_states}
\begin{tabularx}{\linewidth}{
>{\raggedright\arraybackslash}p{0.22\linewidth}
>{\centering\arraybackslash}p{0.12\linewidth}
>{\raggedright\arraybackslash}X}
\toprule
\textbf{Base environment} & \textbf{Dim.} & \textbf{Oracle state variables} \\
\midrule
TwoRoom
& 2D
& Agent position \((x,y)\). \\

Push-T
& 10D
& Goal position \((x,y)\), goal orientation \((\sin\theta,\cos\theta)\), block position \((x,y)\), block orientation \((\sin\theta,\cos\theta)\), and agent position \((x,y)\). \\

Cube
& 15D
& End-effector position \((x,y,z)\), end-effector yaw \((\cos\theta,\sin\theta)\), gripper opening, gripper contact, gripper velocity, block position \((x,y,z)\), and block quaternion \((w,x,y,z)\). \\

Manipulator Insert-Peg
& 33D
& Arm joint positions, touch sensors, hand pose \((x,z,q_w,q_y)\), peg pose \((x,z,q_w,q_y)\), and target pose \((x,z,q_w,q_y)\). \\
\bottomrule
\end{tabularx}
\end{table}

\subsubsection{Model Architecture Details}
\label{app:model-architecture-details}

For LeWM~\citep{lewm} and Sensorimotor World-Model~\citep{sensori-wm}, we use the authors' official implementations. For DINO-WM~\citep{dinowm}, we use the Stable-WorldModel~\citep{swm} implementation without proprioceptive inputs. For DriftWorld~\citep{driftworld}, we use the authors' implementation with two minor modifications: we integrate frame skip into the action-conditioning interface and add a pixel-space cost function for MPC planning, without changing the main architecture. We use the Push-T model configuration for TwoRoom and Push-T, and the Robomimic model configuration for the Cube and Manipulator environments. For IRASim~\citep{irasim}, we use the authors' implementation with the same frame-skip integration and pixel-space planning cost, and use IRASim-XL/2 for all environments.

For the oracle-state experiments, we use three toy predictor families derived from the corresponding end-to-end modeling motifs. Toy-AdaLN keeps the LeWM transformer size unchanged, replacing only the pixel-latent input and output with true-state projections. Toy Action-Duplicate preserves the DINO-WM-style action duplication, causal masking, and shifted prediction design, but adapts it to scalar true-state tokens with a smaller 192D predictor instead of the original high-dimensional patch-token predictor. Toy-DriftWorld preserves the stochastic particle-generation objective and drifting loss, but replaces the image U-Net with a compact vector-valued residual FiLM MLP operating on oracle states.

\section{Training Details}
\label{app:training-details}

\begin{table}[t]
\centering
\small
\setlength{\tabcolsep}{8pt}
\renewcommand{\arraystretch}{1.12}
\caption{Training details and selected checkpoints for end-to-end world models. }
\label{tab:e2e_training_details}
\begin{tabular}{lllll}
\toprule
\textbf{Dataset} & \textbf{Model} & \textbf{LR} & \textbf{Configured duration} & \textbf{Selected checkpoint} \\
\midrule
Push-T & LeWM & - & - & official LeWM checkpoint \\
Push-T & Sensorimotor-WM & \(1\times10^{-4}\) & 10 epochs & epoch 10 \\
Push-T & DINO-WM & - & - & official DINO-WM checkpoint \\
Push-T & DriftWorld & \(5\times10^{-4}\) & max 500k steps & step 500k \\
Push-T & IRASim & \(5\times10^{-5}\) & max 300k steps & step 300k \\
\midrule
Cube & LeWM & - & - & official LeWM checkpoint \\
Cube & Sensorimotor-WM & \(1\times10^{-4}\) & 10 epochs & epoch 10 \\
Cube & DINO-WM & - & - & official DINO-WM checkpoint \\
Cube & DriftWorld & \(5\times10^{-5}\) & 300-epoch cap & step 100k \\
Cube & IRASim & \(5\times10^{-5}\) & max 300k steps & step 300k \\
\midrule
TwoRoom & LeWM &- & - & official LeWM checkpoint \\
TwoRoom & Sensorimotor-WM & \(1\times10^{-4}\) & 10 epochs & epoch 10 \\
TwoRoom & DINO-WM & - & - & official DINO-WM checkpoint \\
TwoRoom & DriftWorld & \(1\times10^{-4}\) & 300-epoch cap & step 500k \\
TwoRoom & IRASim & \(5\times10^{-5}\) & max 300k steps & step 300k \\
\midrule
Insert-Peg & LeWM & \(5\times10^{-5}\) & 10 epochs & epoch 10 \\
Insert-Peg & Sensorimotor-WM & \(1\times10^{-4}\) & 10 epochs & epoch 10 \\
Insert-Peg & DINO-WM & \(5\times10^{-4}\) & 10 epochs & epoch 10 \\
Insert-Peg & DriftWorld & \(5\times10^{-5}\) & max 100k steps & step 100k \\
Insert-Peg & IRASim & \(5\times10^{-5}\) & max 300k steps & step 300k \\

\bottomrule
\end{tabular}
\end{table}
\begin{table}[t]
\centering
\small
\setlength{\tabcolsep}{6pt}
\renewcommand{\arraystretch}{1.12}
\caption{Training details for the selected oracle-state predictor checkpoints.}
\label{tab:oracle_predictor_training}
\begin{tabular}{lllll}
\toprule
\textbf{Dataset} & \textbf{Model} & \textbf{selected LR} & \textbf{Configured duration} & \textbf{Selected checkpoint} \\
\midrule
Push-T & Toy-AdaLN & \(3\times10^{-4}\) & 100 epochs & epoch 100 \\
Push-T & Toy-Action-Duplicate & \(3\times10^{-4}\) & 100 epochs & epoch 100 \\
Push-T & Toy-DriftWorld & \(1\times10^{-4}\) & 50 epochs, max 500k steps & step 500k \\
\midrule
Cube  & Toy-AdaLN & \(3\times10^{-4}\) & 100 epochs & epoch 100 \\
Cube  & Toy-Action-Duplicate & \(3\times10^{-4}\) & 100 epochs & epoch 100 \\
Cube  & Toy-DriftWorld & \(1\times10^{-4}\) & 50 epochs, max 500k steps & step 500k \\
\midrule
TwoRoom  & Toy-AdaLN & \(3\times10^{-4}\) & 100 epochs & epoch 100 \\
TwoRoom  & Toy-Action-Duplicate & \(3\times10^{-4}\) & 100 epochs & epoch 100 \\
TwoRoom  & Toy-DriftWorld & \(1\times10^{-4}\) & 50 epochs, max 500k steps & epoch 50 \\
\midrule
Insert-Peg & Toy-AdaLN & \(3\times10^{-4}\) & 100 epochs & epoch 100 \\
Insert-Peg & Toy-Action-Duplicate & \(3\times10^{-4}\) & 100 epochs & epoch 100 \\
Insert-Peg & Toy-DriftWorld & \(1\times10^{-4}\) & 50 epochs, max 500k steps & epoch 50 \\
\bottomrule
\end{tabular}
\end{table}

AdaLN predictors and Action duplicated predictors use AdamW, batch size \(256\), weight decay \(10^{-3}\), linear warmup followed by cosine decay, gradient clipping at \(1.0\), and an MSE loss over all shifted state predictions. Toy-DriftWorld is trained with AdamW, batch size \(128\), learning rate \(10^{-4}\), weight decay \(10^{-2}\), a \(500\)-step warmup followed by a constant schedule, gradient clipping at \(2.0\), and the drifting particle loss. For end-to-end world models, we follow the original settings.

\section{Detailed Planning and Evaluation Configuration}
\label{app:evaluation}

All planning experiments use the same MPC evaluation pipeline based on the
Stable World Model library~\cite{swm}, with Cross-Entropy Method (CEM) solver.  

\subsection{Planning Evaluation Criteria}
\label{app:planning-success-criteria}

\begin{table}[h]
\centering
\small
\caption{Planning success criteria. Goals are defined by the dataset future
state 25 raw steps after the initial state. An episode is successful if the
criterion is satisfied at least once within 50 environment steps.}
\label{tab:planning-success-criteria}
\begin{tabular}{p{0.16\linewidth} p{0.78\linewidth}}
\toprule
Environment & Planning success criterion \\
\midrule
TwoRoom &
The 2D Euclidean distance between the agent position and the goal agent position
is less than 16 pixels. Target and door positions are not used. \\

PushT &
The 4D Euclidean error over agent and block positions is less than 20 pixels,
and the block angle error is less than $\pi/9$ radians. The angle error accounts
for object symmetries. \\

Cube &
The rollout must simultaneously satisfy: end-effector 3D position error
$\le 0.04$ m, end-effector yaw error $\le 0.10$ rad, and cube 3D position error
$\le 0.04$ m. Cube orientation, gripper opening, contact, and velocity are not
used. \\

Insert Peg &
The rollout must simultaneously satisfy: hand 2D position error $\le 0.04$ m,
hand planar angle error $\le 0.10$ rad, peg 2D position error $\le 0.04$ m, and
peg planar angle error $\le 0.10$ rad. Arm joints, touch, velocity, and the
physical target or slot pose are not used. \\
\bottomrule
\end{tabular}
\end{table}
Table~\ref{tab:planning-success-criteria} summarizes the success criteria used
for each environment. Some details are below.

\paragraph{TwoRoom.}
For TwoRoom, the goal is the dataset future agent position. A rollout is
successful when $\|p_t-p_g\|_2<16$, where $p_t$ is the current agent position
and $p_g$ is the goal agent position. Distances are measured in pixels in the
$224\times224$ environment coordinate system. Target and door positions are not
used.

\paragraph{PushT.}
For PushT, the position error is computed over the 4D vector $[\text{agent}_x,\text{agent}_y,\text{block}_x,\text{block}_y]$. A rollout is successful when both $\|s_t^{\mathrm{pos}}-s_g^{\mathrm{pos}}\|_2<20$ and $d_{\mathrm{angle}}(\theta_t,\theta_g)<\pi/9$ hold at the same timestep. The angle distance accounts for object symmetries: square and \texttt{+} shapes use
$90^\circ$ symmetry, \texttt{Z} and \texttt{I} shapes use $180^\circ$ symmetry, and other shapes use $360^\circ$.

\paragraph{Cube.}
For Cube, the goal is defined by the dataset future end-effector pose and cube  position. A rollout is successful when
$\|p_t^{ee}-p_g^{ee}\|_2\le 0.04\,\mathrm{m}$,
$|\operatorname{wrap}(\psi_g^{ee}-\psi_t^{ee})|\le 0.10\,\mathrm{rad}$, and
$\|p_t^{cube}-p_g^{cube}\|_2\le 0.04\,\mathrm{m}$ all hold at the same timestep. Cube orientation, gripper opening, contact, and velocity are not used. Once satisfied, episode-level success remains true. Cube rollouts are not terminated early by this criterion.

\paragraph{Manipulator Insert Peg.}
For Manipulator Insert Peg, the goal is defined by the dataset future hand and peg poses. Positions are measured in the $(x,z)$ plane, and planar angles are computed from quaternion components $(q_w,q_y)$ as
$2\operatorname{atan2}(q_y,q_w)$. A rollout is successful when the hand position, hand angle, peg position, and peg angle errors are all within their thresholds: $0.04\,\mathrm{m}$ for positions and $0.10\,\mathrm{rad}$ for angles. Arm joints, touch sensors, velocity, and the physical target or slot pose are not used. This dataset-goal success metric differs from the native task success based on the DeepMind Control shaped reward.

\subsection{Configuration}
\label{app:evaluation-configuration}

We evaluated all true-state dynamics models using the same goal-conditioned receding-horizon controller. For each episode, the initial state and goal were sampled from the same dataset trajectory with a temporal offset of 25 environment steps. Planning was performed in the normalized action space using the cross-entropy method (CEM). Each candidate consisted of five action blocks, where each block contained five consecutive environment actions, yielding an effective planning horizon of 25 environment steps. At every planning round, CEM evaluated 300 candidate sequences for 30 iterations and refitted a factorized Gaussian distribution to the 30 lowest-cost elites. The objective was the squared Euclidean distance between the terminal predicted state and the goal in the model’s normalized state space. The final mean sequence was executed open-loop for all five action blocks before replanning. Episodes were evaluated for at most 50 environment steps, corresponding to at most two planning rounds. 

\section{Results}
\label{app:results}
\subsection{Full Results}
\label{app:full-results}
This section reports the full numerical results used in the main analysis. We include absolute MPC planning success rates complementing the relative comparisons shown in the main text, as well as table of relative prediction errors with absolute base-environment rollout errors. For the five-step prediction error, corresponding to \(25\) environment steps, Avoid-T is omitted because its episodes are too short to support this horizon.

\begin{table}[h!]
\centering
\small
\setlength{\tabcolsep}{0.7pt}
\renewcommand{\arraystretch}{0.6}
\caption{Absolute base-task rollout MSE at 5- and 25-step prediction horizons.}
\label{tab:full_next_error}
\begin{tabular}{@{}llcccccccc@{}}
\toprule
\textbf{Base-Env}                     & \textbf{Horizon} & \textbf{LeWM}     & \textbf{SMWM}     & \textbf{DINO-WM}  & \textbf{DriftWorld} & \textbf{IRASim}   & \textbf{AdaLN}      & \textbf{Action Dup.} & \textbf{Toy-Drift}      \\ \midrule
\multirow{2}{*}{TwoRoom}    & 5       & 0.065188 & 0.000058 & 0.033841 & 0.000154   & 0.003838 & 0.00000466 & 0.0000144   & 0.000051   \\ \cmidrule(l){2-10} 
                            & 25      & 0.511843 & 0.000754 & 0.23277  & 0.002164   & 0.023646 & 0.000145   & 0.000636    & 0.00111671 \\ \midrule
\multirow{2}{*}{PushT}      & 5       & 0.00743  & 0.02933  & 0.05124  & 0.001331   & 0.005813 & 0.000005   & 0.000028    & 0.000066   \\ \cmidrule(l){2-10} 
                            & 25      & 0.16739  & 0.23039  & 0.48649  & 0.018599   & 0.050163 & 0.000229   & 0.001068    & 0.00388751 \\ \midrule
\multirow{2}{*}{Insert Peg} & 5       & 0.000717 & 0.002913 & 0.172737 & 0.001448   & 0.00939  & 0.003526   & 0.009356    & 0.054486   \\ \cmidrule(l){2-10} 
                            & 25      & 0.010678 & 0.024729 & 1.410328 & 0.013113   & 0.0593   & 0.053553   & 0.149251    & 0.404549   \\ \midrule
\multirow{2}{*}{Cube}       & 5       & 0.00431  & 0.00203  & 0.15249  & 0.00095    & 0.01027  & 0.000056   & 0.000045    & 0.0521936  \\ \cmidrule(l){2-10} 
                            & 25      & 0.07875  & 0.0193   & 0.89757  & 0.011023   & 0.072107 & 0.000486   & 0.000364    & 0.458214   \\ \bottomrule
\end{tabular}
\end{table}

\begin{table}[h!]
\centering
\small
\setlength{\tabcolsep}{0.9pt}
\renewcommand{\arraystretch}{0.6}
\caption{Base-relative one-step (5 environment step) prediction error. Values report the error ratio with respect to the corresponding base task.}
\label{tab:full_next_error}
\begin{tabular}{@{}llcccccccc@{}}
\toprule
& \multicolumn{6}{c}{End-to-End World Models}
& \multicolumn{3}{c}{Oracle State Predictors} \\
\midrule
\textbf{Env} & \textbf{Variant}
& \textbf{LeWM}
& \textbf{SMWM}
& \textbf{DINO-WM}
& \textbf{DriftWorld}
& \textbf{IRASim}
& \textbf{AdaLN}
& \textbf{Action Dup.}
& \textbf{Toy-Drift} \\
\midrule

TwoRoom
& TR-Route
& 1.08
& 0.27
& 0.97
& 0.01
&
0.90& 0.05
& 1.03
& 0.01 \\
\midrule

& AvoidT
& 1.42
& 1.59
& 1.14
& 0.71
& 0.77
& 1.19
& 1.28
& 1.32 \\
\cmidrule(l){2-10}

& StackT
& 8.01
& 2.12
& 2.38
& 2.54
&
1.36& 475.61
& 115.57
& 70.45 \\
\cmidrule(l){2-10}

\multirow{-3}{*}{PushT}
& Anti-goal
& 4.79
& 1.91
& 2.06
& 1.64
&
1.09& 51.95
& 13.07
& 20.02 \\
\midrule

& Avoid-Peg
& 6.83
& 1.88
& 1.42
& 3.18
&
1.01& 1.45
& 0.62
& 0.08 \\
\cmidrule(l){2-10}

\multirow{-2}{*}{Insert Peg}
& Extract-Peg
& 4.43
& 2.99
& 1.31
& 2.59
&
1.09& 53.53
& 20.01
& 3.55 \\
\midrule

& CubeHop
& 22.53
& 83.97
& 1.46
& 1.38
&
1.19& 13.96
& 30.12
& 0.06 \\
\cmidrule(l){2-10}

\multirow{-2}{*}{Cube}
& Cube Diagonal
& 1.70
& 5.83
& 1.12
& 1.09
&
0.97& 940.34
& 1026.48
& 0.58 \\
\bottomrule
\end{tabular}
\end{table}

\begin{table}[h!]
\centering
\small
\setlength{\tabcolsep}{0.9pt}
\renewcommand{\arraystretch}{0.6}
\caption{Base-relative five-step (25 environment step) prediction error. Values report the error ratio with respect to the corresponding base task.}
\label{tab:full_25_error}
\begin{tabular}{@{}llcccccccc@{}}
\toprule
& \multicolumn{6}{c}{end-to-end world models}
& \multicolumn{3}{c}{oracle state predictors} \\
\midrule
\textbf{Env}
& \textbf{Variant}
& \textbf{LeWM}
& \textbf{SMWM}
& \textbf{DINO-WM}
& \textbf{DriftWorld}
& \textbf{IRASim}
& \textbf{AdaLN}
& \textbf{Action Dup.}
& \textbf{Toy-Drift} \\
\midrule

TwoRoom
& TR-Route
& 0.94
& 0.19
& 0.76
& 0.01
&
0.82& 0.06
& 0.82
& 0.01 \\
\midrule

& StackT
& 5.56
& 2.33
& 2.22
& 2.34
&
1.44& 158.79
& 47.59
& 22.09 \\
\cmidrule(l){2-10}

\multirow{-2}{*}{PushT}
& Anti-goal
& 4.27
& 2.09
& 1.91
& 1.72
&
1.28& 30.03
& 12.69
& 8.96 \\
\midrule

& Avoid-Peg
& 10.54
& 5.52
& 2.10
& 2.94
&
1.06& 2.97
& 1.34
& 0.36 \\
\cmidrule(l){2-10}

\multirow{-2}{*}{Insert Peg}
& Extract-Peg
& 4.08
& 7.40
& 1.74
& 2.75
&
1.19& 34.39
& 12.52
& 4.49 \\
\midrule

& CubeHop
& 21.92
& 103.72
& 1.75
& 1.45
&
1.37& 32.24
& 181.40
& 0.10 \\
\cmidrule(l){2-10}

\multirow{-2}{*}{Cube}
& Cube Diagonal
& 1.54
& 11.32
& 1.12
& 1.05
&
0.97& 841.69
& 1202.22
& 0.46 \\
\bottomrule
\end{tabular}
\end{table}

\begin{table}[h!]
\centering
\small
\setlength{\tabcolsep}{2.7pt}
\renewcommand{\arraystretch}{0.6}
\caption{Absolute planning success rate (\%) across different models and tasks.}
\label{tab:raw_plan}
\begin{tabular}{@{}lcccccccc@{}}
\toprule
\textbf{Task} & \textbf{LeWM} & \textbf{SMWM} & \textbf{DINO-WM} & \textbf{DriftWorld} & \textbf{IRASim} & \textbf{AdaLN} & \textbf{Action Dup.} & \textbf{Drift Toy} \\
\midrule

\rowcolor{chunksky}
TwoRoom & 89.30 & 100.00 & 100.00 & 90.00 & 76.00 & 100.00 & 99.30 & 100.00 \\
\rowcolor{chunksky}
TR-Route & 80.67 & 100.00 & 100.00 & 73.33 & 62.00 & 100.00 & 100.00 & 100.00 \\
\midrule

Push-T & 84.70 & 84.70 & 74.00 & 43.30 & 5.30 & 92.70 & 92.00 & 89.30 \\
AvoidT & 27.33 & 85.33 & 11.33 & 0.00 & 0.00& 100.00 & 99.33 & 99.33 \\
StackT & 46.00 & 32.00 & 40.67 & 19.33 & 4.00& 70.67 & 67.33 & 54.00 \\
Push-T Antigoal & 56.67 & 50.67 & 42.00 & 14.00 & 5.33& 81.33 & 80.67 & 76.00 \\
\midrule

\rowcolor{chunksky}
Insert-Peg & 26.67 & 70.00 & 35.33 & 63.33 & 2.43 & 74.00 & 71.33 & 71.33 \\
\rowcolor{chunksky}
Avoid-Peg & 28.00 & 74.00 & 37.33 & 68.67 & 35.33& 92.00 & 89.33 & 80.67 \\
\rowcolor{chunksky}
Extract-Peg & 6.00 & 24.00 & 11.33 & 18.67 & 9.33& 16.67 & 25.33 & 19.33 \\
\midrule

Cube & 30.70 & 72.70 & 60.00 & 30.70 & 10.70 & 30.00 & 20.70 & 25.30 \\
Cube-Hop & 7.33 & 75.33 & 88.67 & 48.67 & 6.00 & 95.33 & 58.00 & 78.67 \\
Cube-Diagonal & 27.33 & 52.00 & 55.33 & 34.67 & 7.33 & 18.67 & 11.33 & 18.67 \\
\bottomrule
\end{tabular}
\end{table}
\subsection{Factor Support versus Observation Support}
\label{app:factor_vs_obs_support}

Our compositional shifts are defined at the level of physical factors: test
queries recombine factor values that are individually observed in the training
data. However, this does not imply that the resulting observations are
in-distribution for an end-to-end encoder.

Let $\mathcal{Z}^{\mathrm{tr}}$ denote the joint support of physical factors
under the training trajectories, and let
$\mathcal{Z}^{\times}=\prod_{k=1}^K \mathcal{Z}^{\mathrm{tr}}_k$ denote the
Cartesian product of their marginal supports. A compositional query may satisfy
$\mathcal{Z}^{\mathrm{te}}\subseteq \mathcal{Z}^{\times}$ while still containing
factor combinations absent from $\mathcal{Z}^{\mathrm{tr}}$.

End-to-end world models observe $s=f(z)$, where $f$ is the rendering map. Thus
the corresponding observation supports are
$\mathcal{S}^{\mathrm{tr}}=f(\mathcal{Z}^{\mathrm{tr}})$ and
$\mathcal{S}^{\times}=f(\mathcal{Z}^{\times})$. When $f$ mixes factors, unseen
factor combinations in $\mathcal{Z}^{\times}\setminus\mathcal{Z}^{\mathrm{tr}}$
can induce observations in
$\mathcal{S}^{\times}\setminus\mathcal{S}^{\mathrm{tr}}$.

Therefore, a query can be factor-wise in support while still creating an
observation-level shift for an end-to-end encoder. This is why we separately
analyze oracle-state predictors: they remove the encoder-inference problem and
isolate whether the learned dynamics model can generalize across recombined
state-action factors.

\subsection{Additional Observations}
\label{app:additional-observations}
\paragraph{Prediction Space Affects Metric Interpretation}

We do not use the rollout results to directly rank JEPA-style and generative world models. These models predict in different spaces: JEPA-style models are evaluated in latent space, whereas generative models are evaluated in pixel or reconstruction space. As a result, absolute prediction errors are not directly comparable across model families, and we primarily interpret rollout degradation relative to each model's own base-task performance.

Under this relative comparison, generative models often show smaller rollout degradation under some goal shifts, while several JEPA-style models show larger increases. However, this tendency does not translate consistently to planning: a model can show stable rollout error but poor MPC success, or larger rollout degradation while remaining competitive in control. We therefore treat prediction-space effects as an interpretation caveat rather than a model-family ranking. The results suggest that rollout robustness and downstream control utility are related but distinct properties.

\paragraph{Oracle state access does not guarantee better planning.}

Interestingly, access to the oracle state does not always lead to stronger
planning performance. In Cube, several end-to-end models achieve higher
planning success than their oracle-state counterparts, both on the base task
and on CubeDiagonal. This should not be interpreted as evidence that learned
latents are generally superior to physical state representations, since the
end-to-end and oracle predictors differ in more than just their input space.

Rather, these results suggest that state observability and planning suitability
are distinct properties. Oracle states preserve the underlying physical factors,
but the resulting geometry may not be the most favorable one for the planner's
cost and optimization procedure. Conversely, a learned representation may, in
some cases, emphasize task-relevant variation or compress nuisance factors in a
way that makes goal-directed optimization easier. We therefore treat planning
performance as depending not only on how much state information is available,
but also on how that information is organized in the representation space.

\section{Toward Compositional Generalization in World Models}
\label{app:future_compgen_theory}

Our experiments empirically show that removing visual abstraction is not sufficient for reliable goal-shift generalization: free-space dynamics often transfer across new state--action compositions, whereas contact-rich dynamics remain substantially harder even when the predictor receives task-relevant physical state directly. This raises a dynamics-level question: what structure
must a world model learn for its dynamics to remain reusable under compositional
shifts?

Ideally, we want to solve AnyGoal by learning reusable dynamics that extrapolate
to the combinatorial space. That is, we want a learned transition function
\(\hat T\) that matches the observed training transitions on
\(\mathcal C^{\mathrm{tr}}\), while also predicting correctly on novel
compositions in
\(\mathcal C^{\times}\setminus \mathcal C^{\mathrm{tr}}\). We separate this
question into two parts: identifying reusable mechanisms on the training support,
and extrapolating those mechanisms to novel compositions.

\begin{tcolorbox}[notebox]
\textbf{Note.}
The discussion below should be read as a conditional statement, not a claim that reusable mechanisms are identifiable from expert rollouts alone. Rather, we give
one possible sufficient condition: if sparse regime-conditioned local mechanisms are identified on training support and test goals query only locally covered
active inputs, then the model can generalize to globally novel state--action
compositions.
\end{tcolorbox}

\subsection{Identifying Reusable Mechanisms on the Training Support}
\label{app:identifying_r}

To discuss identifiability, we first need to specify what structure the true
transition may have. One observation from our results is that world models often
face multiple \textit{regimes}, such as free motion, contact, grasping, or collision.

In a single-regime setting with little interaction, such as free-space navigation
of an agent, one may expect a transition of the form $z_{t+1}=f(z_t)+g(a_t)$.
In this favorable case, identifiability and compositional generalization are
easier to obtain, for example through an additive function class~\citep{dong2022first, lachapelle2023additive, wiedemer2024compositional, wiedemer2024provable, mahajan2025compositional}. 

However, world-model dynamics are often regime-dependent and more complex than a purely additive and separated form. The action effect is usually better written as \(g_r(z_t,a_t)\) rather than \(g(a_t)\), meaning that the transition cannot be decomposed into simple independent state-only and action-only terms. A simple additive form is therefore too limited for the setting we want to characterize.

We instead consider a factorized transition with regime-dependent local mechanisms. For each factor \(k\), let
\[
z_{k,t+1}=T_{k, r_t}(u_{k,t}),
\qquad
u_{k,t}
=
\bigl(z_{\mathrm{pa}(k),t},a_{\mathrm{pa}(k),t}\bigr),
\]
where \(\mathrm{pa}(k)\) denotes the local parents of factor \(k\), and
\(r_t\in\{1,\ldots,R\}\) denotes a discrete local regime. The transition kernel is specified as follows
\[
T_{r_t}(z_t,a_t)
=
\bigl(
T_{1, r_t}(u_{1,t}),
\ldots,
T_{K, r_t}(u_{K,t})
\bigr).
\]

Since multiple regime--mechanism decompositions can explain the same training transitions, we next state the favorable assumptions under which the discussion applies: a Markov, goal-invariant oracle state, and training-support identification of the relevant local mechanisms up to observational equivalence.

\begin{assumption}[Markov and invariant oracle state]
The oracle state \(z_t\) is Markov, and the transition law is invariant across
training and test goals:
\[
p(z_{t+1}\mid z_{\le t},a_{\le t})
=
p(z_{t+1}\mid z_t,a_t),
\qquad
T_{\mathrm{tr}}(z_{t+1}\mid z_t,a_t)
=
T_{\mathrm{te}}(z_{t+1}\mid z_t,a_t).
\]
Thus, goal shifts change the behavior distribution, but not the environment
dynamics.
\end{assumption}



\begin{assumption}  
[Identification of local mechanisms]
For each state factor $k$, suppose that the interaction regime $r_t$ and the
local input $u_{k,t}=u_k(z_t,a_t)$ are observed or identifiable from the
training transitions, and that the corresponding regime can be consistently inferred at prediction time.
\[
    z_{k,t+1}=T_{k,r_t}(u_{k,t}),
    \qquad
    \hat z_{k,t+1}=\hat T_{k,r_t}(u_{k,t}).
\]
For each factor and regime, define the training support
\[
    \mathcal U^{\mathrm{tr}}_{k,r}
    :=
    \bigcup_{t=0}^{H-1} \operatorname{supp}_{p_{\mathrm{tr}}}
    (U_{k,t}\mid R_t=r).
\]

Let $\mathcal L_{\mathrm{tr}}(\tilde T)$ denote the population objective used
to train the world model. We assume that this objective identifies the
regime-conditioned mechanisms on their training support: any population
minimizer $\tilde T$ satisfies
\[
    \tilde T_{k,r}(u)=T_{k,r}(u),
    \qquad
    \forall u\in\mathcal U^{\mathrm{tr}}_{k,r},
\]
for every $k,r$, up to observational equivalence and regime relabeling.

For a model $\hat T$ learned from finite data, we allow estimation and
optimization error
\[
    \epsilon_{k,r}
    :=
    \sup_{u\in\mathcal U^{\mathrm{tr}}_{k,r}}
    \|\hat T_{k,r}(u)-T_{k,r}(u)\|_2,
\]
and assume $\epsilon_{k,r}\rightarrow 0$ as the amount of relevant training
data grows and the training objective is optimized consistently.
\end{assumption}


\subsection{Extrapolating Identified Mechanisms to Novel Compositions}
\label{app:extrapolating_i}

Training-support identification does not by itself imply compositional
generalization. Two mechanisms can agree on all observed training transitions and
still disagree on a novel state--action composition induced by a goal shift.
Recombination therefore requires additional structure that separates coordinates
that affect a mechanism from coordinates that are only part of the global state.

Additive or separable functions provide one favorable case, which allow us to reuse the learned mechanisms for each factor $z_{k, t+1}$, but contact-rich
world-model dynamics are often regime-dependent rather than purely additive. We
propose a simpler scenario where each transition mechanisms $T_{k, r_t}$ depends sparsely on the inputs $s_t, a_t$. For a fixed regime $r$, let
\(x_t=(z_t,a_t)\), then
\[
z_{k,t+1}=T_{k, r_t}(x_t).
\]
Sparsity of \(\nabla_x T_{k, r_t}(x)\) implies that only a subset of coordinates affects
the \(k\)-th factor transition, while the remaining coordinates may change the
global composition without changing this local mechanism.

\begin{assumption}[Sparse local mechanisms]
For each factor $k$ and regime $r$, there exists a fixed coordinate set
$P_{k}\subseteq\{1,\ldots,\dim(x)\}$, with
$|P_{k}|\ll \dim(x)$, such that
\[
    T_{k,r}(x)=T_{k,r}(x_{P_{k}}).
\]
Equivalently,
\[
    \frac{\partial T_{k,r}(x)}{\partial x_j}=0
    \qquad \forall j\notin P_{k}.
\]
We write $u_{k,t}=x_{P_{k}, t}$ for the inputs relevant to the
$k$-th transition mechanism.
\end{assumption}

Sparse local mechanism can allow us to create favorable cases when a goal shift creates a novel \emph{global} composition without creating novel \emph{local} mechanism inputs.

\begin{assumption}[Local support coverage]
Let
\[
\mathcal U_{k,r}^{\mathrm{te}}
:=
\bigcup_{t=0}^{H-1}
\operatorname{supp}_{p_{\mathrm{te}}}(U_{k,t}\mid R_t=r)
\]
denote the test support of the inputs of transition mechanism for factor \(k\). For every factor
\(k\), assume
\[
\mathcal U_{k,r}^{\mathrm{te}}
\subseteq
\mathcal U_{k,r}^{\mathrm{tr}}.
\]
Thus, the global state--action composition may be novel, but every local
mechanism is queried only on local inputs covered during training.
\end{assumption}


\begin{tcolorbox}[notebox]
\textbf{Intuition.}
Within a fixed regime, sparse local support separates active \& inactive
coordinates. The active coordinates \({u_{k}}\) determine the transition mechanism for the factor $z_k$, while coordinates outside \(u_{k}\) can be recombined to create
globally novel state--action compositions without changing that local mechanism.
Thus, the sparser the mechanism, the larger the set of globally novel
compositions that can be covered by the same local transition rule.
\end{tcolorbox}



\begin{proposition}[Local active-coordinate coverage controls global error]
\label{prop:local_coverage_error} 
Let $\mathcal L_{\mathrm{tr}}(\tilde T)$ denote the learning objective used to train the world model. 
Suppose that we perfectly optimize the world model on training support, then under assumption 1 and 2, we identify the learned model such that the true transition and learned transition share the same regime-conditioned
local decomposition,
\[
T(z_t,a_t)
=
\bigl(
T_{1,r_t}(u_{k ,t}),
\ldots,
T_{K,r_t}(u_{k ,t})
\bigr),
\]
and
\[
\hat T(z_t,a_t)
=
\bigl(
\hat T_{1,r_t}(u_{k ,t}),
\ldots,
\hat T_{K,r_t}(u_{k ,t})
\bigr).
\]
Assume that, for every factor \(k\) and regime \(r\), we satisfy the assumption of sparse local mechanisms (assumption 3) and local support coverage (assumption 4).

Then, for any test composition \((z_t,a_t)\in\mathcal C^{\mathrm{te}}\),
\[
\left\|
\hat T(z_t,a_t)-T(z_t,a_t)
\right\|_2
\le
\left(
\sum_{k=1}^{K}\epsilon_{k,r_t}^2
\right)^{1/2}.
\]
where
\[
\epsilon_{k,r}
:=
\sup_{u\in\mathcal U_k^{\mathrm{tr}}}
\left\|
\hat T_{k,r}(u)-T_{k,r}(u)
\right\|_2 .
\]

Consequently, if \(\epsilon_{k,r}\to0\) for all \(k,r\), then the world model
generalizes to novel combinations present in the test support.
\end{proposition}

\paragraph{Proof.}
Take any test composition $(z_t,a_t)\in\mathcal C^{\mathrm{te}}$ with interaction regime $r_t$.
By Assumption~4, for every factor $k$,
\[
u_{k,t}\in\mathcal U_{k,r_t}^{\mathrm{te}}
\subseteq
\mathcal U_{k,r_t}^{\mathrm{tr}}.
\]
Thus, each local mechanism is evaluated within its regime-conditioned training support. By Assumption~2,
\[
\left\|
\hat T_{k,r_t}(u_{k,t})
-
T_{k,r_t}(u_{k,t})
\right\|_2
\le
\epsilon_{k,r_t}.
\]
Since the true and learned transitions share the same factorized decomposition,
\[
\begin{aligned}
\left\|
\hat T(z_t,a_t)-T(z_t,a_t)
\right\|_2^2
&=
\sum_{k=1}^{K}
\left\|
\hat T_{k,r_t}(u_{k,t})
-
T_{k,r_t}(u_{k,t})
\right\|_2^2 \\
&\le
\sum_{k=1}^{K}\epsilon_{k,r_t}^2.
\end{aligned}
\]
Taking the square root completes the proof.

\paragraph{Discussion.}
This construction should be read as an illustrative sufficient condition rather
than a complete solution. Sparse local mechanisms reduce a high-dimensional
global coverage problem to smaller local coverage problems, but they do not by
themselves solve mechanism identification, regime discovery, or extrapolation to
genuinely new local interactions. The broader challenge is to develop world-model
architectures, objectives, and data-collection strategies that identify reusable
action-conditioned mechanisms and impose enough structure on their interactions
to support generalization beyond expert rollouts.

\section{AnyGoal Library and Reproducibility}
\label{app:anygoal_library}

We release AnyGoal as an open-source Python library for collecting
goal-shifted datasets, training world models, and evaluating rollout prediction
and MPC planning. The library provides both Python APIs and a unified CLI,
allowing researchers to reproduce the benchmark experiments and evaluate
additional models under the same protocol. AnyGoal builds on
Stable-WorldModel~\citep{swm} and retains the original model
implementations and numerical protocols.
Complete installation instructions and API references are available in the
\href{https://hazel-heejeong-nam.github.io/anygoal/docs/}{online documentation}.

\subsection{Installation and Dataset Access}
\label{app:installation_a}

The library can be installed with
evaluation dependencies using \texttt{uv}.

\begin{lstlisting}[style=anygoal]
pip install uv
uv venv --python 3.10
source .venv/bin/activate
uv pip install "anygoal[eval]"
\end{lstlisting}

The default installation provides dataset access, collection, validation,
and simulation environments, while the \texttt{eval} and \texttt{all}
extras add evaluation and training dependencies, respectively.

The CLI exposes a registry of supported datasets and models.

\begin{lstlisting}[style=anygoal]
# Inspect available datasets and models
anygoal list datasets
anygoal list models

# Inspect default data and checkpoint paths
anygoal config path data
anygoal config path checkpoints
\end{lstlisting}

Datasets and pretrained checkpoints can be downloaded from Hugging Face
through the artifact interface.

\begin{lstlisting}[style=anygoal]
# Download training and test datasets
anygoal artifacts download --dataset pusht
anygoal artifacts download --dataset avoidt

# Download a pretrained model checkpoint
anygoal artifacts download --model lewm
\end{lstlisting}

Model integrations use the corresponding upstream implementations.
Their source code can be installed separately.

\begin{lstlisting}[style=anygoal]
anygoal models list
anygoal models install lewm
\end{lstlisting}

\paragraph{Dataset Collection}

Beyond the released evaluation episodes, AnyGoal provides tools for
generating additional goal-shifted trajectories. Dataset collection is
available through both Python and CLI interfaces, which use the same
in-package collectors.

The following example collects 400 Avoid-T episodes using eight
parallel environments.

\begin{lstlisting}[style=anygoal,language=Python]
import anygoal

dataset = anygoal.collect(
    "avoidt",
    num_episodes=400,
    num_parallel=8,
    seed=0,
)
\end{lstlisting}

The equivalent collection procedure can be invoked from the terminal.

\begin{lstlisting}[style=anygoal]
anygoal collect avoidt --num-episodes 400 --num-parallel 8

# Collect another goal-shifted variant
anygoal collect cubediagonal --output /path/to/cubediagonal.h5
\end{lstlisting}

\subsection{World-Model Evaluation}
\label{app:world-model_e}

AnyGoal provides standardized interfaces for evaluating a base-trained
world model on both its original task and goal-shifted variants.
The same model checkpoint can be used for rollout prediction and
downstream MPC planning, enabling comparisons across evaluation metrics.

\paragraph{Planning evaluation.}
The following commands evaluate LeWM on Push-T and Avoid-T.

\begin{lstlisting}[style=anygoal]
# Original training goal
anygoal eval planning --model lewm  --dataset pusht --seed 0

# Unseen goal
anygoal eval planning --model lewm  --dataset avoidt --seed 0
\end{lstlisting}

\paragraph{Rollout evaluation.}
The corresponding rollout-prediction evaluation uses the same
model and dataset identifiers.

\begin{lstlisting}[style=anygoal]
anygoal eval rollout --model lewm  --dataset pusht --seed 0

anygoal eval rollout --model lewm  --dataset avoidt --seed 0
\end{lstlisting}

By default, the evaluation commands resolve the checkpoint associated
with the selected model and base-task family through the shared artifact
catalog. Users can also specify a checkpoint explicitly with
\texttt{--checkpoint} or download missing artifacts with
\texttt{--download}.

The evaluation interfaces additionally support explicit random seeds,
allowing comparisons across repeated runs.

\subsection{Reproducibility and Extensibility}
\label{app:reproducibility}

AnyGoal organizes datasets, checkpoints, and supported model
implementations through shared registries. Dataset and model identifiers,
artifact locations, and evaluation configurations can be specified
explicitly, reducing ambiguity when reproducing experiments.
The library exposes its core functionality through Python APIs,
including \texttt{anygoal.collect}, \texttt{anygoal.train},
\texttt{anygoal.evaluate\_rollout}, and \texttt{anygoal.plan}.
These interfaces complement the CLI and support integration into
existing research workflows.

The released benchmark contains 400 episodes per goal-shifted
variant, while the collection tools allow researchers to construct
additional evaluation datasets. Together with the standardized
evaluation interfaces, this enables future world models to be tested
on the same controlled goal shifts without reproducing the
environment and evaluation infrastructure.
The source code, configuration files, and implementation instructions
are available at
\href{https://github.com/galilai-group/AnyGoal}{github.com/galilai-group/AnyGoal}.

\end{document}